\documentclass[fleqn,10pt]{wlscirep}
\usepackage[utf8]{inputenc}
\usepackage[T1]{fontenc}
\usepackage{lineno}
\usepackage{float}
\usepackage{graphicx}%
\usepackage{multirow}%
\usepackage{amsmath,amssymb,amsfonts}%
\usepackage{amsthm}%
\usepackage{mathrsfs}%
\usepackage[title]{appendix}%
\usepackage{xcolor}%
\usepackage{textcomp}%
\usepackage{manyfoot}%
\usepackage{booktabs}%
\usepackage{algorithm}%
\usepackage{algorithmicx}%
\usepackage{algpseudocode}%
\usepackage{listings}

\title{Active Sensing Shapes Real-World Decision-Making through Dynamic Evidence Accumulation}

\author
[1,$\dag$]{Hongliang Lu}
\author[2,3,$\dag$]{Yunmeng Liu}
\author[3]{Junjie Yang}
\affil[1]{The Hong Kong University of Science and Technology, Civil and Environmental Engineering, Hong Kong, 999077, China}
\affil[2]{The Hong Kong University of Science and Technology, Emerging Interdisciplinary Area, Hong Kong, 999077, China}
\affil[3]{The Hong Kong University of Science and Technology (Guangzhou), Systems Hub, Guangzhou, 511453, China}
\affil[$\dag$]{these authors contributed equally to this work}

\begin{abstract}
Human decision-making heavily relies on \textbf{active sensing}, a well-documented cognitive behaviour for evidence gathering to accommodate ever-changing environments. However, its operational mechanism in the real world remains non-trivial. Currently, an in-laboratory paradigm, called evidence accumulation modelling (EAM), points out that human decision-making involves transforming external evidence into internal mental beliefs. However, the gap in evidence affordance between real-world contexts and laboratory settings hinders the effective application of EAM. Here we generalize EAM to the real world and conduct analysis in real-world driving scenarios. A cognitive scheme is proposed to formalize real-world evidence affordance and capture active sensing through eye movements. Empirically, our scheme can plausibly portray the accumulation of drivers' mental beliefs, explaining how active sensing transforms evidence into mental beliefs from the perspective of information utility. Also, our results demonstrate a negative correlation between evidence affordance and attention recruited by individuals, revealing how human drivers adapt their evidence-collection patterns across various contexts. Moreover, we reveal the positive influence of evidence affordance and attention distribution on decision-making propensity. In a nutshell, our computational scheme generalizes EAM to real-world contexts and provides a comprehensive account of how active sensing underlies real-world decision-making, unveiling multifactorial, integrated characteristics in real-world decision-making.

\textbf{Keywords:} Decision Making, Evidence Accumulation Modelling, Active Sensing, Real-world Driving

\end{abstract}

\begin{document}

\flushbottom
\maketitle
%  Click the title above to edit the author's information and abstract

\thispagestyle{empty}

%\noindent Please note: Abbreviations should be introduced at the first mention in the main text – no abbreviations lists or tables should be included. Structure of the main text is provided below.

\section*{Introduction}\label{sec1}

%% 第一段
Far from being a passive recipient, humans can actively gather and identify the information valuable for them (namely, evidence) to reinforce the mental belief for their intended behavioural decision-making \cite{thomas2019gaze,gottlieb2012attention,gottlieb2018towards}; such a cognitive characteristic can be conceptualized as ‘\textbf{active sensing}’ \cite{yang2016theoretical}, formally describing an active evidence-collection behaviour inherent in humans. 
In such an ever-changing, itinerant world, active sensing comes into play moment-to-moment, enabling humans to gauge their decisions and nudge them into profitable situations.
%Real-world driving is indeed the case,
%in which active sensing is commonplace and well-characterized, appearing as the deliberate collection of evidence to manage ensuing decision-making consequences \cite{vellenga2022driver,blakemore2001perception}.
% 举例
As a case in point, active sensing is commonplace and well-characterized in real-world driving, appearing as drivers' deliberate collection of evidence to manage ensuing decision-making consequences \cite{vellenga2022driver,blakemore2001perception}. 
%A driver intending to change lanes will actively embark on attending to the real-time manoeuvre of the nearest front and rear vehicles on the target lane; only when it is deemed reachable enough will the lane-changing behaviour be executed \cite{xia2021human,han2019driving}.
Simply defined, active sensing can inform humans to render them more nuanced and rational to make sense of their decisions in the real world; yet to date, we still fall short of understanding how active sensing shapes human mental beliefs in decision-making and how it leads to the ultimate decision in real-world contexts.

%% 第二段
% 尽管决策边界理论的范畴很好的解释了人们在决策中的信念积累
%Recent years have witnessed the growing promise of a well-documented in-lab theory, evidence accumulation modelling (EAM), in plausibly underwriting human mental processing. 
In recent years, evidence accumulation modelling (EAM), a well-documented theory from numerous in-lab research, has shown increasing potential in explaining mental processing through the transition from (external) evidence to (inner) mental belief.
Specifically, given a behavioural intention, relevant decision-making evidence continues to accumulate into a mental belief until it crosses a bound, at which point the corresponding decision is ultimately formed \cite{glickman2022evidence,hoxha2023accounting,van2019relation}. 
%%%%%%%%%%%%
Previous studies about EAM have demonstrated its theoretical value; however, their experimental contexts are typically in-lab or computer-simulated, and the generalization of EAM theory in the real world remains challenging for two major reasons.
First, in feature-rich scenarios (for example, selection between the same commodities with different prices, or job offers with different salaries), the mental belief can be captured handily as humans can glean informative evidence, with the evidence affordance being explicit and overt enough (like price and salary) \cite{kiverstein2024experience,jones2003affordance,gibson1977theory}. Rather, the real-world context is always a feature-poor scenario in which the evidence affordance is no longer straightforward to quantify, understanding the role of active sensing in shaping mental beliefs through EAM will become significantly intricate and intractable.
Second, the evidence-collection behaviour does not always follow an immutable pattern and could vary across different contexts \cite{sepulveda2020visual,shevlin2021attention}, even when evidence affordance is quantifiable. Different evidence affordances may lead to different patterns of evidence collection (for example, a congested road may recruit drivers' more attention to collect decision-making evidence, whereas it is quite the opposite on a clear road; in turn, evidence collection determines what evidence afforded is leveraged by them) \cite{callaway2021fixation,rangelov2020evidence}.
% 总结
Together, a parsimonious description of real-world contexts and a comprehensive understanding of the interplay between evidence affordance and evidence collection are imperative to causally understand the evidence accumulation occurring in the real world.

% replication: resulted in statistically significant results in the same direction as the original work.  
%Establishing the same pattern of behavior across different implementations may endow the replicated finding with more credibility - and an enlarged prospect of real-world relevance.   Replication studies that not only closely reproduce an existing version of a study setup and analysis, but in addition test its generalizability to different versions (e.g., of a task) or new contexts (e.g., real-life interventions for lab-based studies) have a tremendous potential to deepen our understanding of what past results mean.
% “Establishing the same pattern of behavior across different implementations may endow the replicated finding with more credibility—and an enlarged prospect of real-world relevance.”

% 第三段讲做了啥
In light of EAM, here we provide a cognitively plausible scheme that captures how active sensing works in practice in real-world contexts, called the dynamic evidence accumulation model (DEAM).
We capitalize on perceptual states, a fuzzy description of feature-poor items, to formalize the external information of real-world driving scenarios in a tractable fashion, and two external evidence affordances (external factors) for decision-making can be developed: \textbf{evidence bias}, which reflects whether the current scenario is profitable to make a certain decision, and \textbf{evidence clarity}, which indicates whether the current scenario is easy to make any decision.
We first dissect active sensing in real-world decision-making through the lens of human eye movements \cite{gottlieb2014attention,orquin2013attention,ahlstrom2021eye}, by which the self-evidence-collection pattern of humans (i.e., \textbf{attention}, standing as internal factors) can be taken into account in our analysis. 
% 总结(生理和心理因素)
By doing this, we can understand how active sensing leads to real-world decision-making from the perspectives of both external and internal factors.

% 第四段讲贡献
Based on our scheme, a real-world driving dataset along with the records of drivers' eye movements is used for our analysis \cite{alletto2016dr,palazzi2018predicting}.
All findings in our study can be tracked following Fig.\ref{fig1}c.
First, we furnish a computational description of how active sensing determines evidence information utility \cite{harkins1987information,stewart2022humans}, shapes drivers' mental beliefs plausibly, and leads to ultimate decisions (Fig.\ref{fig1}a). 
%%%%%%%%%%
Our computational scheme exemplifies the generalization of previous in-lab findings to real-world contexts.
% Our computational scheme demonstrates the potential for the generalization of in-lab theory to real-world contexts with adjustments
%
Next, we demonstrate a negative correlation between evidence clarity and attention (reaction time and attentional switching) recruited by individuals, unveiling the interplay between (external) evidence affordance and (self) evidence collection: humans adapt their evidence collection based on evidence affordance, and in turn, their evidence collection determines which evidence can be afforded.
In closing, we highlight a strikingly positive correlation between decision-making propensity and corresponding evidence affordance, as well as the consolidating effect of attention distribution on decision-making propensity, demonstrating that real-world driving decision-making is a multifactorial, integrated result of the external world and the self-evidence-collection behaviour \cite{gluth2020value}.
To conclude, we reveal the prowess of active sensing in explaining the way human drivers engage with the external environment during decision-making, allowing for a better understanding of how individuals actively absorb external information and transform it into their inner representations in real life.

\begin{figure}[!h]
\centering
\includegraphics[width=0.65\textwidth]{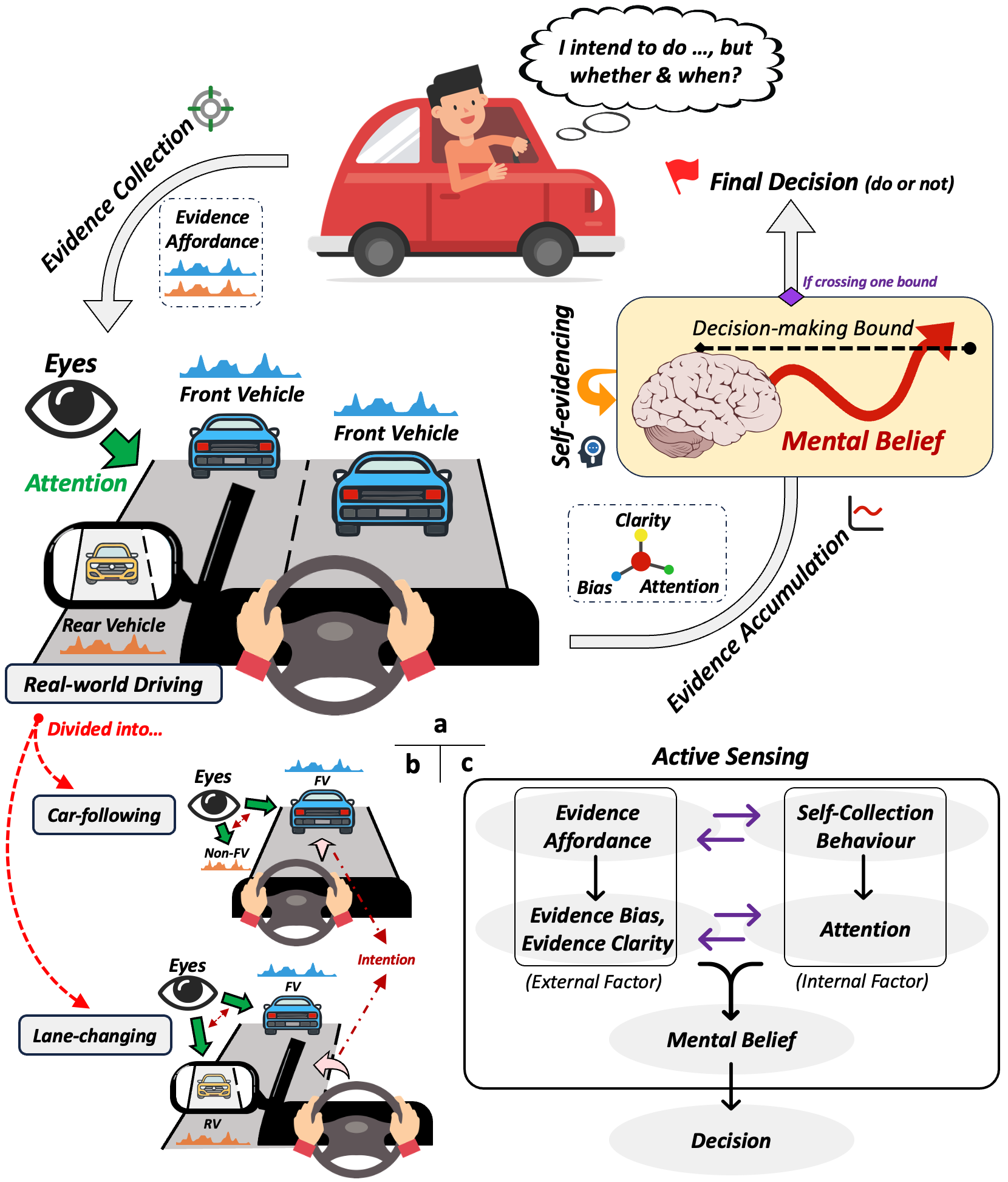}
\caption{Overall formalization.
    (a) Illustration of how active sensing comes into play in real-world driving.
(b) Two types of real-world driving scenarios: lane-changing and car-following.
(c) An overview of our findings.
}\label{fig1}
\end{figure}

\section*{Results}\label{sec2}

% 总
Decision-making in real-world driving mainly encompasses two types: lane-changing and car-following. Figure \ref{fig1}b gives an overall look. 
% 简单介绍lane-changing
As for the former, drivers intending to change lanes will actively monitor the nearest front and rear vehicles (FV and RV) in the target lane to ensure enough reachability for lane-changing. 
% 简单介绍car-following
For the latter, drivers decide whether to accelerate or decelerate to maintain a subjectively acceptable distance from the FV in the current lane. 
% 总结
Both of them entail active sensing before decision-making, so we will delve into how active sensing comes into play within these two scenarios. 
% 介绍数据集
A real-world open-source dataset, DR(eye)VE \cite{alletto2016dr,palazzi2018predicting}, is used for analysis in our study, containing over 370 minutes and 500,000 frames of real-world driving data along with the records of real-time drivers' eye movements. We identify and extract all latent lane-changing and car-following events and feature them with quantitative evidence affordance (detailed in Section \ref{sec4}). 
% 潜在事件的含义
Note that latent events refer to those situations where drivers intend to change lanes or follow other vehicles, indicated by precursors like eye movements \cite{wedel2023modeling,zhou2009cognitive}, but may not actually do so.

\subsection*{Drivers' Mental Beliefs Portrayed by DEAM}\label{subsec2.1}

Here we show the mental belief accumulation of human drivers portrayed by DEAM in real-world decision-making. We borrow from the distribution of momentary evidence, which is used to investigate the regulatory role of attention in evidence collection \cite{jang2021optimal}. Thus, how humans adapt their evidence-collection patterns across diverse driving contexts can be understood from the perspective of the utility of evidence information, i.e., information richness and effectiveness (see Section \ref{sec4} for methodological details). Based on momentary evidence, DEAM can portray mental belief as a dynamic computation of relative decision value (RDV).

\subsubsection*{Lane-changing} 
\begin{figure}[H]
\centering
\includegraphics[width=0.75\textwidth]{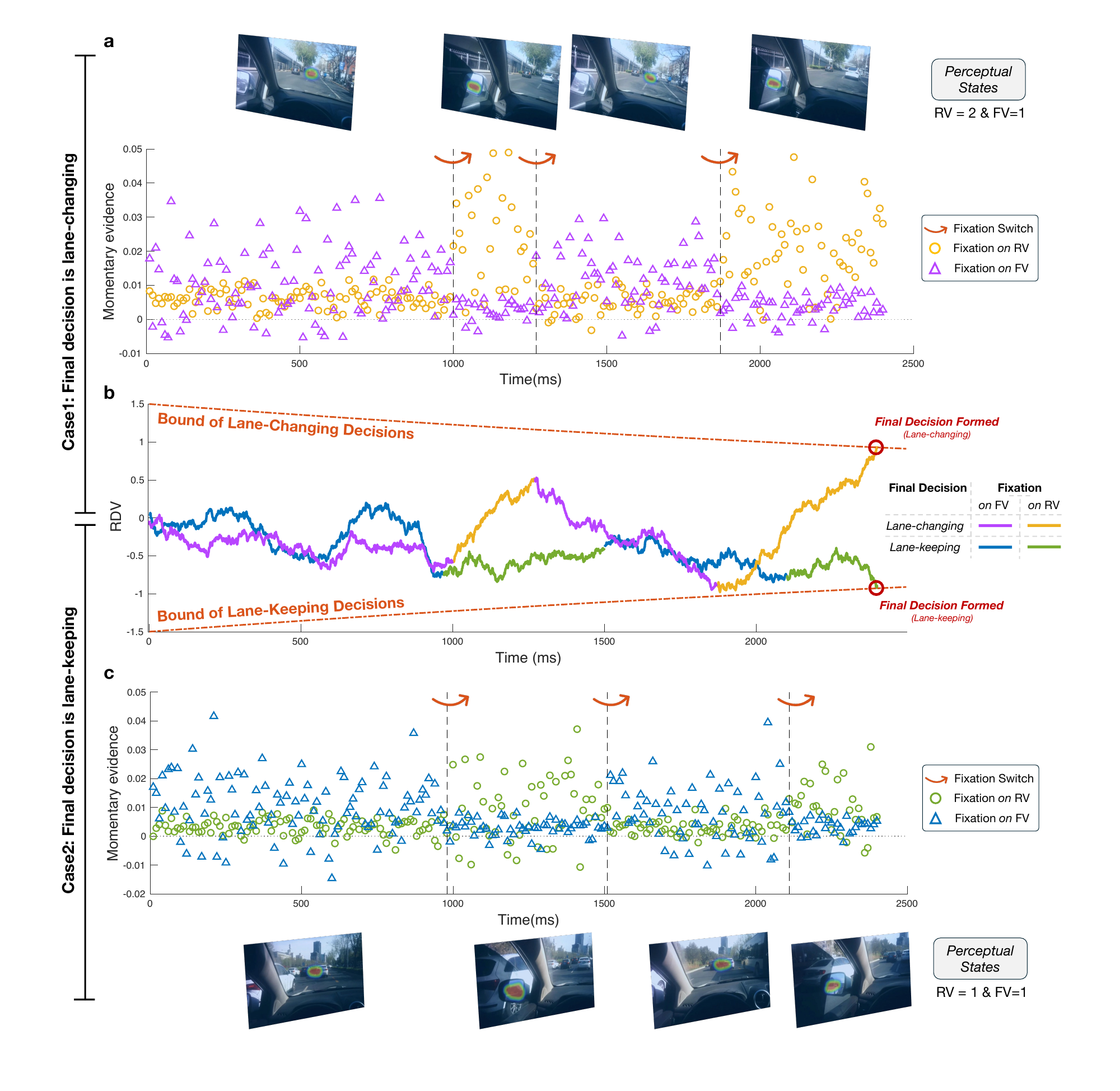}
\caption{The mental beliefs of drivers in different lane-changing scenarios portrayed by DEAM. 
(a) First-person view scenarios and distribution of momentary evidence for a lane-changing decision portrayed by DEAM. 
We plot the distribution of momentary evidence portrayed by DEAM, received by drivers at each time step related to lane-changing decision-making. The purple triangular scatters represent the momentary evidence the driver receives from FV. The yellow circular scatters represent the momentary evidence from the RV. %When the driver's attention is focused on one item, that item will provide better evidence utility, characterized by a wider distribution and higher mean of its momentary evidence. 
For this descriptive figure, we use the following parameters: $\sigma^2_z = 1, \theta = 0.3, \Delta t = 0.01$. 
To better illustrate the distribution of the driver's attention, we provide first-person view images of scenarios that human drivers receive in the upper panel and the heatmap representing the spatial distribution of the driver's visual attention during that period. %The upper panel shows the first-person view scenario of the decision-making process in which the driver's final decision is lane-changing. 
(b) Dynamical computation of RDV signals over time portrayed by DEAM. 
The blue RDV curves and purple lines represent the driver's attention being allocated to the FV, while the yellow and green represent the attention allocated to the RV. 
The orange dashed lines in the figure represent the decision bounds of the DEAM, while the red circles denote the points at which the decision-making process is concluded. When the accumulated mental belief reaches the upper bound, it is interpreted as the driver having made the decision for lane-changing. Conversely, reaching the lower bound indicates lane-keeping decisions.
For this descriptive figure, we use the following parameters: $d = 0.002$, $\sigma^2_\text{model} = 0.02^2$, $\theta = 0.3$, $\Delta t=0.001$.
(c)First-person view scenarios and distribution of momentary evidence for a lane-keeping decision portrayed by DEAM.  The blue triangular scatters represent the momentary evidence the driver receives from FV. The green circular scatters represent the momentary evidence from RV. We use the following parameters for this descriptive figure: $\sigma^2_z = 1, \theta = 0.3, \Delta t = 0.01$. The bottom
panel shows the first-person view scenario of the decision-making process in which the driver's final decision is lane-keeping.}
\label{fig2}
\end{figure}

For lane-changing, human drivers engage in decision-making mental belief accumulation by integrating evidence from two available decision candidates: lane-changing and lane-keeping. Figure \ref{fig2} showcases two sample cases with drivers' first-person view images, where Figure \ref{fig2}a is a case with a final decision of lane-changing while \ref{fig2}c is lane-keeping. 
%再讲证据
Figure \ref{fig2}b shows the RDV curves of two sample cases, where the purple and yellow curve indicates Case 1 with the final decision of lane-changing, while the blue and green curve indicates Case 2 of lane-keeping. Figure \ref{fig2}a and \ref{fig2}c show the momentary evidence drivers receive from FV and RV separately during active sensing, sampled at intervals of $\Delta t=0.01$ seconds, where the Gaussian distribution plot on each scenario image corresponds to the item that the driver's visual attention focuses on.

Figure \ref{fig2}a and \ref{fig2}c demonstrate that attention enhances the utility of evidence information. When attention is focused on a particular item (FV or RV), the attended item will provide higher evidence utility, characterized by a broader distribution that indicates greater information richness, as well as a higher mean value of its momentary evidence to suggest higher information effectiveness. In the meantime, the evidence from the unattended item will become more uncertain, reflected by a more narrow distribution and lower mean of its momentary evidence. When the attentional location varies, evidence utility changes accordingly based on whether the item is attended to.
%举例
Taking the time window from 0 to 1,000 milliseconds in the two sample cases as an example, the attention on FV makes the momentary evidence from FV (triangular scatters) more informative than that from RV (circular scatters). Consequently, the distribution of momentary evidence from FV has a higher mean value compared to that of RV. A more refined result with $\Delta t=0.001$ can be found in Appendix Figure S2 to demonstrate the distribution of momentary evidence information more clearly.

%最后讲RDV
In the following, the received momentary evidence can be accumulated into the driver's mental belief for decision-making over time. Figure \ref{fig2}b shows this process through RDV curves. The RDV curve starts with an initial value of 0, indicating the start of mental belief accumulation; its fluctuation provides an estimate towards the relative attractiveness of the two choices at any given moment. When the RDV curve crosses any decision-making bound (two orange dashed lines, the upper is lane-changing decision bound, and the lower is lane-keeping decision bound), a decision will be formed: a crossing at the bound of lane-changing decisions (the upper red cirque) indicates a formed lane-changing decision, while a crossing at the other (the lower red cirque) indicates a formed lane-keeping decision. 
%In the two cases, the RDV curve will gravitate towards the entities of the evidence sources, reflected by an oscillation between different pieces of evidence. 
Furthermore, another fact can be seen: the deterministic change in RDV during each time step, also known as the integration rate, hinges on the disparity in value between the two entities and the attentional location at each time step (see Supplementary Method in Appendix for methodological details). 
%举例
Taking the time window spanning from 1,300 to 1,850 milliseconds in Case 1 as an example: the perceptual state of RV is 2 and FV is 1, indicating a more influential impact of RV than FV on decision-making in principle. However, when the driver's attention is allocated to FV (purple parts), the RDV curve will be primarily influenced by the evidence from FV rather than RV with a higher perceptual state value. Our scheme highlights the comparable significance of both attention and perceptual state during the decision-making process.

\newpage

\subsubsection*{Car-following}
\begin{figure}[H]
\centering
\includegraphics[width=0.95\textwidth]{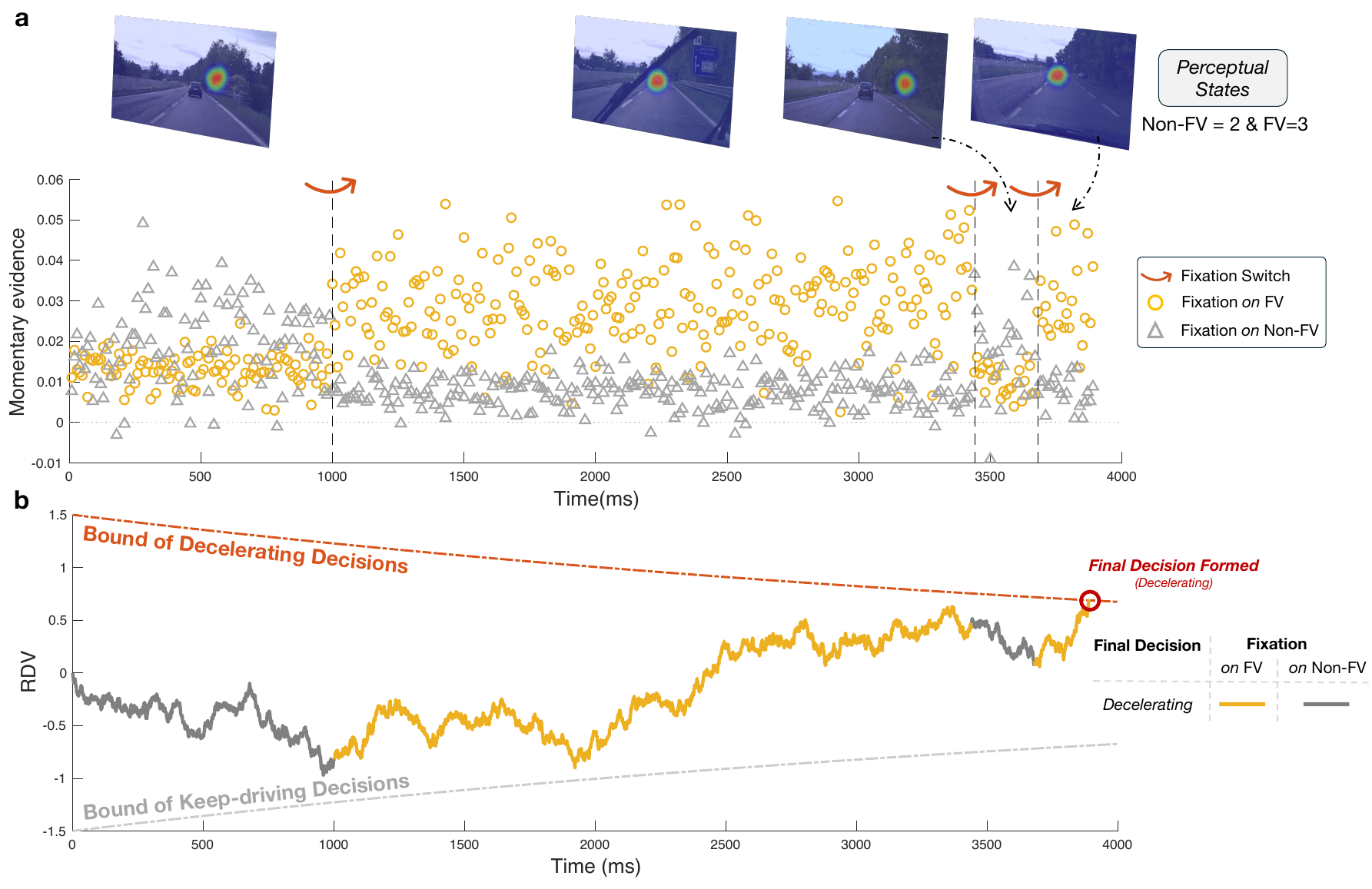}
\caption{The mental beliefs of drivers in car-following scenarios portrayed by DEAM. 
(a) First-person view scenarios and distribution of momentary evidence for a decelerating decision portrayed by DEAM. We present the distribution of momentary evidence received by drivers at each time step for deceleration decision-making, depicted by DEAM. Grey triangular scatters represent momentary evidence from surroundings, while yellow circular markers represent momentary evidence from FV. The parameters used for this descriptive figure are $\sigma^2_z = 1, \theta = 0.4, \Delta t = 0.01$. The upper panel illustrates a first-person view scenario of the decision-making process where the final decision is decelerating. 
(b) Dynamic computation of RDV signals over time modelled by DEAM. The grey RDV curves depict periods when the driver's attention is distributed among surrounding items, while the yellow lines represent periods when attention is allocated to FV. The upper orange dashed line serves as the decision bound for decelerating, with a red circle indicating the point where the decision-making process ends. When the cumulative mental belief reaches the upper bound, it indicates the driver opted for decelerating. Reaching the lower grey bound signifies a decision to maintain the current speed.
For this descriptive illustration, the parameters used are $d = 0.008$, $\sigma^2_\text{model} = 0.02^2$, $\theta = 0.4$, and $\Delta t=0.001$.
}\label{fig3}
\end{figure}

Figure \ref{fig3} demonstrates the results of RDV and the utility of evidence information in a car-following case. Figure \ref{fig3}a displays a set of car-following front-view images and illustrates the momentary evidence distributions received from the FV and surroundings under attention modulation. Consistent with the lane-changing case, the attended item provides better evidence utility. For example, in the 1,000-3,400ms and 3,700-3,900ms time windows, we can infer from the fixation heatmap that the driver's attention focus is on the FV. During these periods, the momentary evidence received by the driver from the FV (yellow circles) exhibits a broader distribution and higher mean, representing greater information richness and higher information effectiveness, respectively. Therefore, the mental belief reflected during these two time periods will drift towards a deceleration decision, with the RDV curve in Figure \ref{fig3}b reflecting the corresponding inner process.

Collectively, we show how DEAM plausibly portrays the mental beliefs of drivers and elucidate this from the perspective of information utility. Our computational scheme provides insight into how human drivers adapt their evidence-collection patterns across diverse driving contexts in real-world decision-making.

\subsection*{From Evidence Affordance to Decision-Making}\label{subsec2.2}

% 证据是如何影响决策

\begin{figure}[H]
\centering
\includegraphics[width=0.67\textwidth]{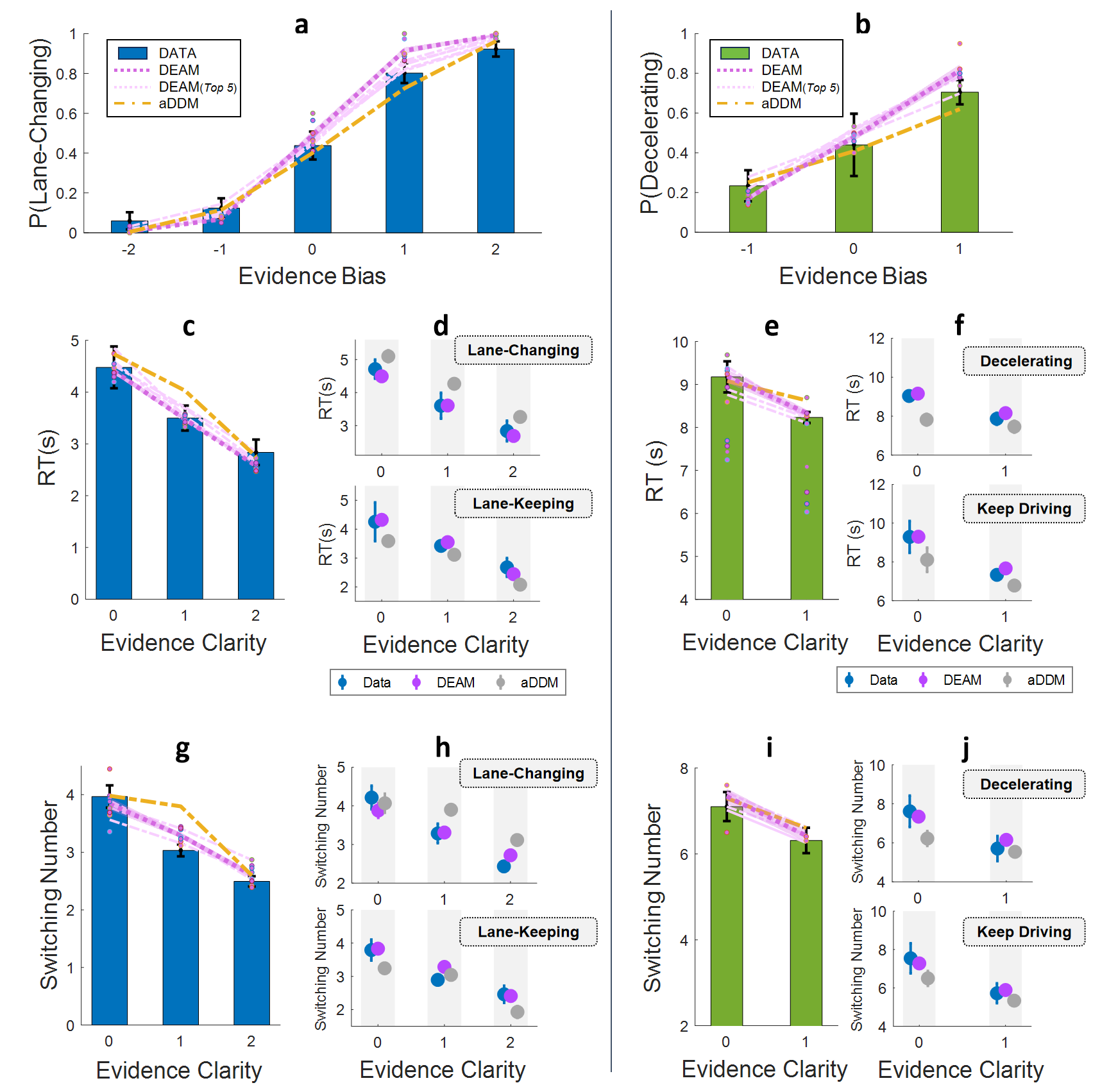}
\caption{Replicating human driver behaviour using DEAM and aDDM in lane-changing and car-following scenarios. 
(a) The probability of lane-changing monotonically increases as a function of evidence bias. The bars represent the empirical data from human driver behaviour ($t(7)=21.8, p<0.001$); the thick pink dashed line represents the fitting results from DEAM ($t(7)=107.4, p<0.001$); the yellow dashed line represents the fitting results from aDDM ($t(7)=72.6, p<0.001$). Additionally, the thin pink dashed lines represent the DEAM fitting results for five other sets of different parameter values. 
(b) Decelerating probability increases consistently with evidence bias (human data: $t(5) = 5.02, p < 0.001$; DEAM: $t(5)=39.32, p<0.001$; aDDM: $t(5) = 33.32, p < 0.001$). 
(c) RT decreases monotonically as a function of evidence clarity (human data: $ t(7)= -3.53, p<0.01$; DEAM: $ t(7)=-8.56, p<0.001$; aDDM: $t(7)= -9.44, p<0.001$). Both human drivers and computational models show shorter RT when more clear evidence for decision-making is provided.  
(d) Comparative analysis of RT and evidence clarity in lane-changing (top) and lane-keeping (bottom) decisions for data, DEAM, and aDDM. %The gap between the mean of RT fitted by DEAM and the mean of actual human driving data is smaller compared to aDDM, reflecting that DEAM has a better ability to reproduce human decision-making in different choices. 
(e) RT in car-following decision-making decreases as evidence clarity increases (human data: $t(5) = 3.03, p < 0.05$; DEAM: $t(5) = -11.82, p < 0.001$; aDDM: $t(5) = -13.20, p < 0.001$). 
(f) Comparative analysis of RT and evidence clarity in decelerating(top) and keep-driving (bottom) decisions for data, DEAM, and aDDM. %
(g) The number of attentional switches in lane-changing decision-making decreases with evidence clarity. For human drivers, the number of attentional switches correlates with evidence clarity ($p<0.001$) and shows a monotonically decreasing trend. For models, decrease in attentional switching number as a function of evidence clarity (DEAM:  $t(7)=-2.4, p<0.05$; aDDM: $t(7)=3.46, p<0.05$). When the evidence for decision-making is clearer, the results show a decrease in the number of attentional switches. 
(h) Comparative analysis of attentional switching number and evidence clarity in lane-changing (top) and lane-keeping (bottom) choices for data, DEAM, and aDDM. %The gap between the mean of switching numbers fitted by DEAM and the mean of actual human driving data is smaller compared to aDDM.
(i) The number of attentional switching in car-following decreases with evidence clarity. For human drivers, the number of attentional switches correlates with evidence clarity ($p < 0.05$). DEAM demonstrates fewer attentional switches with clearer evidence (DEAM: $t(5) = -4.19, p < 0.01$).
(j) Comparative analysis of attentional switching numbers and evidence clarity in decelerating(top) and keep-driving (bottom) decisions for data, DEAM, and aDDM. %DEAM shows a smaller gap between fitted switching numbers and actual human data compared to aDDM.
}\label{fig4}
\end{figure}

%Figure \ref{fig4} demonstrates the results of driving behavioural features recorded in human data and captured by DEAM, across varying levels of decision-making evidence bias and clarity. 
Figure \ref{fig4} demonstrates the results of decision-making behavioural patterns recorded in human ground-truth data and those fitted by DEAM across different driving scenarios. 
%证据对决策的影响
Here, we examine how evidence affordance influences decision-making propensity. 
%and successfully replicate human patterns using DEAM. 

%DEAM: 
%MSE of p(Lane-changing) between simulated and real data: 0.021
%MSE of RT between simulated and real data: 0.71
%MSE of SwitchCount between simulated and real data: 0.19
%ADDM:
%MSE of p(Lane-changing) between simulated and real data: 0.024
%MSE of RT between simulated and real data: 0.84
%MSE of SwitchCount between simulated and real data: 0.36
\subsubsection*{Lane-changing} 
Figure \ref{fig4}a shows the decision-making probability of human ground-truth data, DEAM, and aDDM in lane-changing, under different levels of decision-making evidence bias. 

\textbf{Human:} 
For human patterns, a positive correlation is observed in drivers' ground-truth data ($t(7)=21.8, p<0.001$), indicating that as decision-making evidence bias (ranging from -2 to 2) increases, the evidence affordance provided by the current scenario is more profitable for lane-changing, leading to a higher propensity for lane-changing decisions.

%模型模式-human揭示现象，deam强调一致
\textbf{DEAM:} 
The positive correlation between decision-making propensity and evidence bias is mirrored by both DEAM ($t(7)=107.4, p<0.001$) and aDDM ($t(7)=72.6, p<0.001$).
% 简单介绍mse
Notably, DEAM closely fits human patterns. 
The fitting performance is measured by the Mean Squared Error (MSE), which gauges the discrepancy between a model's predictions and actual outcomes. A lower Mean Squared Error (MSE) indicates that the model more accurately reflects the intrinsic patterns within the data.
%mse对比
In this regard, DEAM demonstrates a strong alignment with human data, with an MSE of approximately 0.021. In comparison, aDDM shows a slightly higher MSE of 0.024, suggesting a slight divergence from the human pattern when compared to DEAM's performance.
%In conclusion, DEAM's superior fit, as evidenced by its lower MSE, reveals its capability to model human patterns in lane-changing decisions under varying levels of evidence bias.
%A lower MSE, as exhibited by DEAM, signifies a model that more accurately reflects human patterns, indicating a better grasp of the factors influencing drivers' lane-changing decisions. 

\subsubsection*{Car-following}
Figure \ref{fig4}b shows the decision-making probability of human ground-truth data, DEAM, and aDDM in car-following, under different levels of decision-making evidence bias. 

\textbf{Human:} 
As decision-making bias escalates within the range of -1 to 1, the evidence affordance provided by the current scenario increasingly favours decelerating, strengthening the propensity towards decelerating decisions (Figure \ref{fig4}b). 

\textbf{DEAM:} 
The statistical significance of this trend is evident across human ground-truth data ($ t(5)= 5.02, p<0.01$), as well as the qualitative replication by DEAM ($t(5)=39.32, p<0.001$) and aDDM ($t(5) = 33.32, p < 0.001$). 
Consistent with the conclusions from lane-changing, DEAM outperforms the aDDM in fitting accuracy (DEAM: MSE=0.60; aDDM: MSE=0.76).

In sum, these findings unveil the positive correlation between decision-making propensity and corresponding evidence affordance and highlight the value of DEAM in capturing human decision-making patterns in driving scenarios, including both lane-changing and car-following contexts.

%\subsection{Impact of Evidence Affordance on Evidence Collection Patterns}\label{subsec2.3}
\subsection*{From Evidence Affordance to Evidence Collection}\label{subsec2.3}
%part1：证据对attention的影响
Here, we analyze how evidence affordance influences drivers' attention patterns, including reaction time (RT) and attentional switching.
% and successfully replicate human behavioural patterns using DEAM. 
\subsubsection*{Lane-changing}
Here we show how evidence affordance influences evidence collection in lane-changing scenarios.

%证据对决策时长的影响
\textbf{Human:} Human ground-truth data reveals a negative correlation between RT and decision-making evidence clarity level (blue bars in Figure \ref{fig4}c; $ t(7)= -3.53, p<0.01$).
This correlation underscores the increased difficulty in decision-making when faced with low evidence clarity. Drivers require longer RT to make decisions when confronted with ambiguous evidence, indicating that more prolonged attention is necessary for the accumulation of mental beliefs in scenarios with challenging decision-making due to unclear evidence. 
%证据对SW的影响
In addition, human patterns reveal the influence of evidence clarity on attentional switching, both in terms of switching numbers and switching probability over time. As shown in Figure \ref{fig4}g, human ground-truth data of lane-changing highlight a correlation between attentional switching number and evidence clarity (Kruskal-Wallis test, $p<0.05$), demonstrating monotonically decreasing trends.
%进一步分析证据是如何影响SW的

\begin{figure}[!h]
\centering
\includegraphics[width=0.9\textwidth]{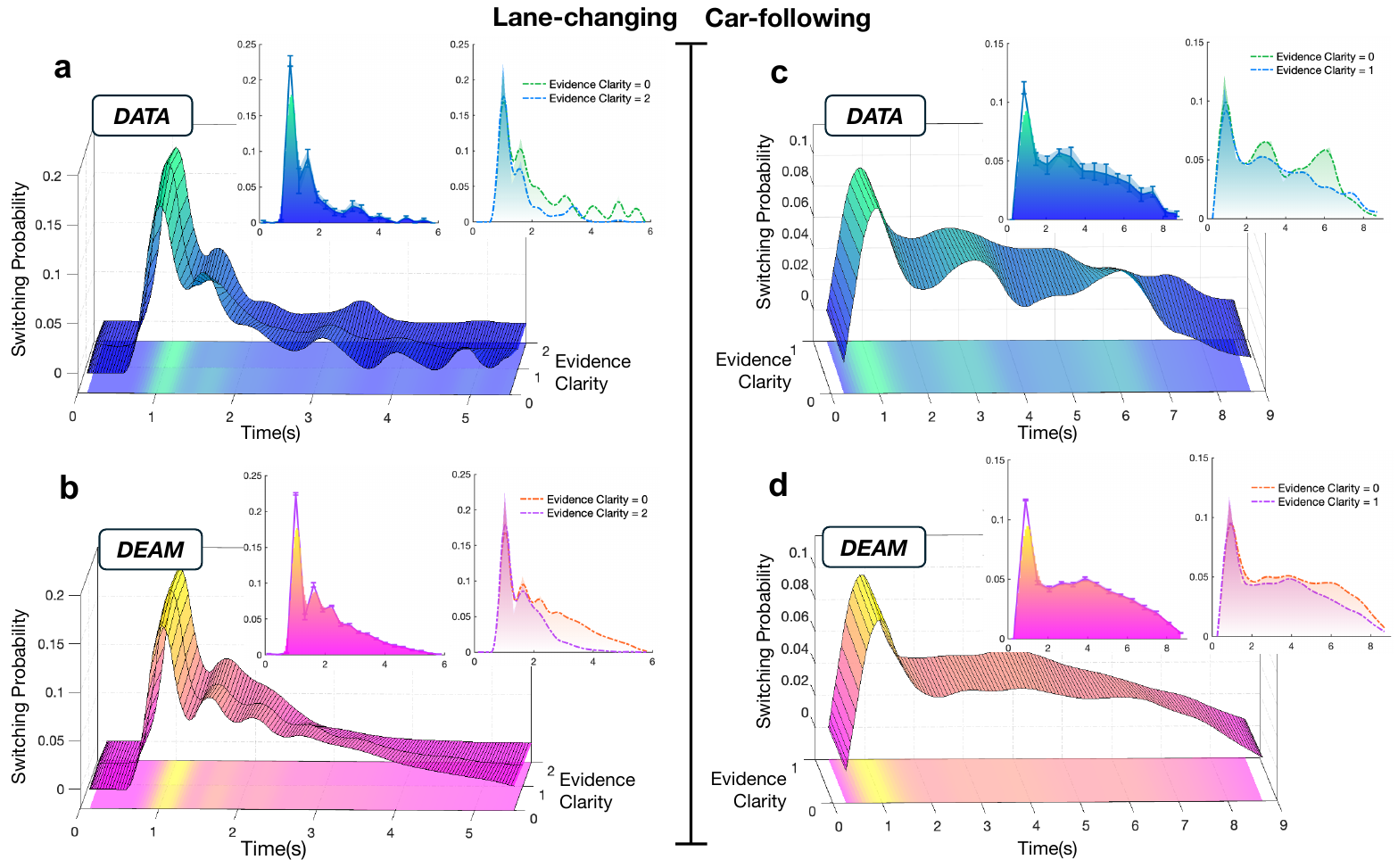}
\caption{Attentional switching probability of human drivers and DEAM across different evidence clarity levels over time. The x-axis in the 3D plot represents elapsed time, the y-axis depicts evidence clarity, and the z-axis represents the smoothed switching probability. The surface is projected on the z=-0.02 plane. 
Insets in the top right corner of the 3d plot separately display the overall attentional switching probability (left insets), and attentional switching probability at the highest and lowest evidence clarity levels (right insets).
(a) Attentional switching probability results of human drivers in lane-changing scenarios. (b) Attentional switching probability results of DEAM in lane-changing scenarios. (c) Attentional switching probability results of human drivers in car-following scenarios. (d) Attentional switching probability results of DEAM in car-following scenarios.}
\label{fig6}
\end{figure}

Figure \ref{fig6} illustrates the variation in attentional switching probabilities over time, providing further insights into how evidence affordance influences attention dynamics. % shedding light on
For lane-changing, Figure \ref{fig6}a shows the overall attentional switching probability for human drivers, in which attentional switching probability peaks at approximately a timestamp of one second (fixation switching from FV to RV). This peak is attributed to the predetermined time window of driver fixations for data processing (detailed in Section \ref{subsec2.2}). At approximately a timestamp of 1.6 seconds, the line reaches the secondary peak (fixation switching from RV to FV). 
The time interval between the first and secondary peaks indicates that after a lane-changing intention is established, the duration of first-stage mental belief accumulating from RV usually lasts approximately 0.6 seconds, namely, the duration of the first fixation on RV.
Next, we partition all the latent lane-changing events according to the level of evidence clarity to compare attentional switching probabilities between challenging (evidence clarity level = 0, represented by the green line in Figure \ref{fig6}a) versus easy scenarios (evidence clarity level = 2, represented by the blue line in Figure \ref{fig6}a). Humans exhibit lower attentional switching probabilities in easier lane-changing scenarios. In other words, drivers do not need to frequently switch their attention between FV and RV to collect more evidence when high levels of evidence clarity are presented.
% level=2切换概率总体偏小--level=2切换次数应该偏小，与前面结论一致
These findings align with the conclusions drawn from Figure \ref{fig4}c, which indicates that drivers tend to demonstrate fewer attentional switching in easier lane-changing scenarios with higher evidence clarity levels. 

\textbf{DEAM:} DEAM effectively captures the human attention patterns mentioned above, including RT, the number of attentional switching, and attentional switching probability over time. 
Firstly, the negative correlation between RT and evidence clarity is captured by both DEAM ($ t(7)=-8.56, p<0.001$) and aDDM ($t(7)= -9.44, p<0.001$).
The models' fitting performance, measured by MSE, shows DEAM with a lower MSE of 0.71 compared to aDDM's 0.84. The lower MSE value reveals DEAM's superior ability to replicate human attentional duration under conditions of varying evidence clarity.
%不同决策下的拟合
In Appendix Figure S3a, changes in RT with evidence clarity in different decisions are depicted for humans, DEAM, and aDDM. 
The superior fitting of DEAM is also reflected across different decision contexts. As shown in Figure \ref{fig4}d, the top panel visualizes RT changes with evidence clarity for lane-changing decisions, while the bottom panel shows the same for lane-keeping decisions. It is evident that DEAM closely aligns with human data, showing minimal deviation regardless of the final decision outcome, especially demonstrating significantly better fitting performance than aDDM in the lane-changing decision context. 

In addition, Figure \ref{fig4}g illustrates the impact of evidence clarity on the attentional switching number, where DEAM demonstrates a similarity with human data, that is, the number of attentional switches decreases with an increase in decision-making evidence clarity level (DEAM: $t(7)=-2.4, p<0.05$). 
The negative correlation between evidence clarity and attentional switching suggests that in situations where decision-making is challenging due to low evidence clarity, additional evidence is required to accumulate mental belief, necessitating multiple attentional switching. 
The fitting performance of the DEAM and aDDM are evaluated using MSE, highlighting DEAM's superior fitting of human data compared to aDDM (DEAM: $MSE=0.19$; aDDM: $MSE=0.36$). DEAM shows significantly better fitting performance than aDDM, particularly at moderate levels of evidence clarity (clarity level=1). In Appendix Figure S3b, further comparison of the fitting across different decisions confirms DEAM's superiority over aDDM. The superiority is particularly prominent in lane-changing decisions, where DEAM exhibits markedly improved fitting capabilities compared to aDDM (Figure \ref{fig4}h). 

Furthermore, DEAM's fitting outcomes of attentional switching probability parallel the human patterns in lane-changing (Figure \ref{fig6}b). DEAM captures the effect of evidence clarity on attentional switching probabilities, showing higher switching probabilities in more challenging scenarios (purple line) compared to easier scenarios (orange line). 
Collectively, these findings underscore DEAM's ability to accurately model how human drivers manage attention resources under varying evidence affordance for decision-making, underscoring its advantage in modelling evidence-collection patterns. 

%%%%%%%%%%%%%%%%%%%%%%%%%%%%%
\subsubsection*{Car-following}
Below, we show how evidence affordance influences evidence collection in car-following scenarios.

%第一段negative关系：总-分结构
\textbf{Human:} The result from human ground-truth data in car-following parallels the earlier lane-changing experiment, demonstrating a negative correlation between evidence clarity and attention (RT and attentional switching) recruited by drivers.
Firstly, Figure \ref{fig4}e shows the negative correlation between RT and evidence clarity ($t(5) = 3.03, p < 0.05$). The green bars indicate that attentional duration increases as evidence clarity diminishes. 
Secondly, Figure \ref{fig4}i presents a correlation between attentional switching number and evidence clarity ( Kruskal-Wallis test; $p<0.05$), with the green bars showing a decreasing trend.
Furthermore, Figure \ref{fig6}c delineates the overall attentional switching probability results for human drivers, where human drivers exhibit higher probabilities of switching attention in more challenging car-following scenarios (evidence clarity = 0, represented by the green line in Figure \ref{fig6}c), compared to easier car-following scenarios (evidence clarity = 1, represented by the blue line in Figure \ref{fig6}c). The negative effect of evidence clarity on attentional switching probability parallels the finding in lane-changing, that is, humans exhibit lower attentional switching probabilities as evidence clarity increases. 

%两个实验不同之处
However, unlike lane-changing scenarios, where attentional switching probabilities significantly decrease over time (e.g., from approximately 9.5\% to 0.5\% in the 1.6s-5s time window), car-following scenarios show a more stable pattern, with attentional switching probabilities hovering around 5\% and displaying a smaller fluctuation across the same time window. Drivers in the car-following scenario tend to maintain a relatively consistent level of attentional switching probabilities over time. 
The distinction in attentional switching patterns across various driving scenarios unravels a behavioural tendency: in situations requiring quick decisions such as lane-changing, drivers tend to promptly engage in early and rapid attentional switching to gather relevant decision-making evidence.

\textbf{DEAM:} The fitting results presented in Figure \ref{fig4}e demonstrate a negative correlation between RT and evidence clarity (DEAM: $t(5)=-11.82, p<0.001$; aDDM: $t(5) = -13.20, p < 0.001$). Both models capture the human patterns that drivers take longer RT when confronted with ambiguous information that could either support decelerating or free-driving. Meanwhile, DEAM exhibits superior fitting performance compared to aDDM (DEAM: $MSE = 0.80$; aDDM: $MSE = 1.49$). Variations in RT across different decisions are illustrated for humans, DEAM, and aDDM, as evidence clarity changes (Figure S4a, Appendix).
In Figure \ref{fig4}f, further examination of the model fitting performance across different decisions reinforces the superiority of the DEAM. DEAM's fitted RT mean is closer to actual human data compared to aDDM, with the contrast being particularly pronounced at low levels of evidence clarity (clarity level = 0), indicating better human decision-making replication across choices. 

%cf
The number of attentional switching exhibited by DEAM also closely aligns with the corresponding human driver data (Figure \ref{fig4}i). DEAM demonstrates a statistically significant negative correlation between the number of attentional switching and the level of evidence clarity ($t(5)=-4.19, p<0.01$). 
Quantitative evaluation of the model fitting performance reinforces the DEAM's superior capacity to capture human decision-making behaviour relative to the aDDM. The DEAM achieves an MSE of 1.58, outperforming the aDDM, which yielded an MSE of 2.11. In Figure \ref{fig4}j, the gap between mean switching numbers fitted by DEAM and those of actual human driving data is smaller compared to aDDM, reflecting that DEAM has a better ability to reproduce human attention patterns across different choices. 

%section 2.3的总结
%Collectively, the results reveal patterns in human data where attention is influenced by evidence clarity across different interactive driving scenarios, with these negative correlations replicated by DEAM.
Collectively, we conducted a comprehensive analysis to understand the impact of evidence clarity on the evidence-collection patterns employed by drivers during real-world driving decision-making. The result demonstrates a negative correlation between evidence clarity and attention (reflected by RT and attentional switching), revealing that humans adapt their evidence collection based on evidence affordance. Our computational model, DEAM, not only replicates these negative correlations but also exhibits a superior capacity to mirror human patterns, providing insights to understand self-evidence-collection behaviour in decision-making.

%\subsection{Consolidating Influence of Evidence Collection on Decision-Making Propensity}\label{subsec2.4}
\subsection*{From Evidence Collection to Decision-Making}\label{subsec2.4}
Furthermore, we examine how attention impacts the propensity for decision-making. Previous in-lab decision-making studies have demonstrated that human choices are swayed by the last visual fixation \cite{krajbich2010visual,jang2021optimal,tavares2017attentional,callaway2021fixation}. 
Echoing these findings, our work captures a parallel behavioural propensity within the context of driving tasks, underscoring the role of attention in steering the course of decision-making and generalizing the patterns to real-world contexts.

\clearpage
\subsubsection*{Lane-changing}
As shown in Figure \ref{fig7}a (upper), across all levels of evidence bias, the probability of lane-changing decisions with final attention on RV consistently surpasses that with final attention on FV. The blue line, representing the lane-changing probability with a final fixation on RV, consistently exceeds the green line, which corresponds to the probability curve with a final fixation on FV.

\subsubsection*{Car-following}
\begin{figure}[H]
\centering
\includegraphics[width=0.9\textwidth]{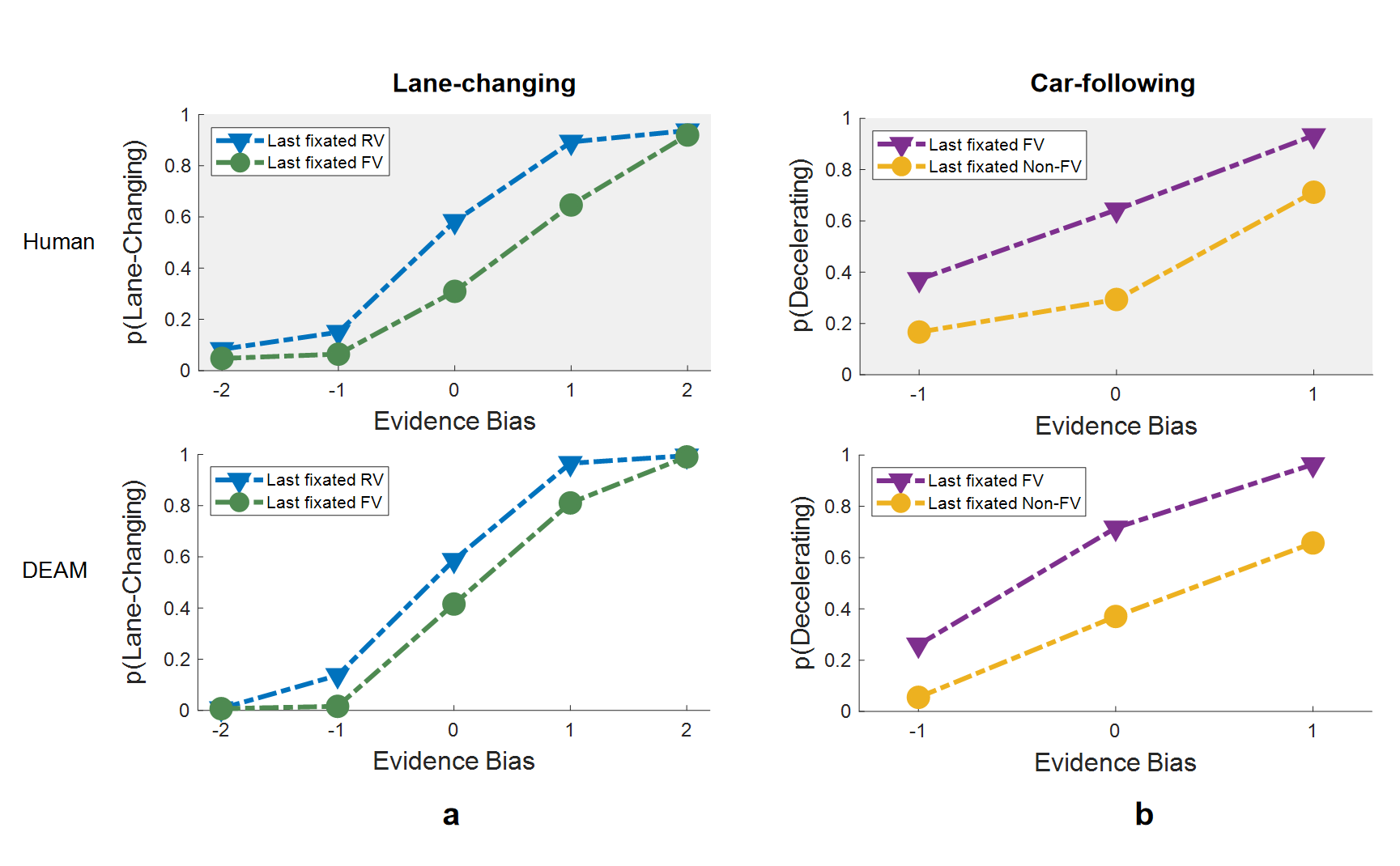}
\caption{
The influence of the last fixation on decision-making captured by human data and DEAM. 
(a) The relationship between lane-changing preference and last fixations across different evidence biases, captured by human driver data (upper) and DEAM (lower). The blue line records the lane-changing probability when the last fixation is on RVs; the green line records the probability when the last fixation is on FVs. 
(b) The relationship between decelerating preference and last fixations across different evidence biases, captured by human driver data (upper) and DEAM (lower). The purple line depicts the probability of deceleration when the last fixation is on FVs, while the yellow line represents the probability when the last fixation was not on FVs. 
}\label{fig7}
\end{figure}

Similarly, in car-following scenarios, the probability of deceleration after lastly fixating on FV consistently surpasses that after lastly driving in a relaxed state and saccading around (as illustrated by the purple line consistently exceeding the yellow line in Figure \ref{fig7}b).

These findings suggest that human drivers' ultimate decisions are notably influenced by their last attention focus, highlighting the consolidating effect of attention distribution on decision-making propensity. To sum up, decision-making is a result of multiple influences, intricately shaped by attention and evidence affordance, as elaborated in Section \ref{subsec2.1}. This inherent hallmark of human behaviour is replicated by DEAM in the lower panel of Figure \ref{fig7}.

\section*{Discussion}\label{sec3}

% 总结一下
In this paper, we delve into how active sensing shapes the mental belief of human drivers in real-world driving based on a cognitively plausible scheme DEAM. The analysis is rolled out from a multifactorial, holistic perspective—(external) evidence affordance and (self) evidence-collection behaviour, concluding with a fact that real-world driving decision-making is an interplay result of the external world and the self behaviour. The empirical results show the promising efficacy of our scheme in causally bridging between intentions and decisions, providing a sound backbone for understanding real-world driving behaviours through the lens of human cognitive behaviours while driving. 
%%%
%%%%%%%%%说一下复制和泛化
Although previous in-lab findings and computational models hold strong theoretical insights into decision-making \cite{krajbich2010visual,jang2021optimal,tavares2017attentional,callaway2021fixation}, 
demonstrating similar behavioural patterns in real-world data strengthens the credibility of these lab-based results \cite{generalization2022}. Our computational framework further exemplifies how such in-lab studies can be effectively generalized to real-world contexts.
% 简单说一下数据局限
Considering the disparity between real-world driving and simulator-based driving, we give priority to the real-world driving data in this study. The prime reason why we select this dataset is because it is currently the only and largest open-source real-world dataset that records both drivers' driving behaviours and eye movements \cite{baee2021medirl}. More real-world data are expected to be considered in future work.

% 我们是task-related
It is worth noting that, as highlighted in \cite{Gottlieb_Oudeyer_Lopes_Baranes_2013}, active sensing \textit{per se} comes into play in two ways: one is task-related, and the other is curiosity-driven. Our work pertains to the former as real-world driving behaviours always follow specific tasks that human drivers intend to undertake, and this is indeed the case this study focuses on. Future studies on the curiosity-driven line may uncover the role of drivers' active sensing in the formation of behavioural intentions \cite{gottlieb2020curiosity}. 
% 我们是value-based
Additionally, our understanding towards drivers' active sensing falls into the scope of value-based decision-making, a well-documented decision-making investigation framework \cite{milosavljevic2010drift,tajima2016optimal}. In this regard, there is a plurality of considerations to concretize values more sensibly, such as subjective values of information \cite{kobayashi2019common, kobayashi2024neural,bobadilla2020subjective}, anticipation \cite{caplin2001psychological}, exploration-exploitation switch \cite{marques2020internal}, and even the combination of them \cite{sharot2020people, kobayashi2019diverse}. Here we capitalize on the theory of descriptive fuzzy logic (detailed in the next section) to circumvent the excessive discussion regarding what factors should be integrated into values, and the result empirically demonstrates that this treatment can work in practice, in the meanwhile maintaining the scalable flexibility for others to factor more elements of interest into our scheme.

% 说一下mental belief的可信性
As for the plausibility of mental beliefs portrayed by DEAM, some relevant cognitive models have been widely used in enormous contemporary visual decision-making tasks and proven capable of fitting human electroencephalogram (EEG) signals well \cite{nunez2017attention,si2020predicting}. In the next section, we detail such plausibility conserved in our scheme. However, ethical and moral issues, as well as the heavy dependence on invasive electrodes for high-fidelity EEG signals at present, prevent us from obtaining EEG signals from real human drivers to validate the consistency of our scheme with real neural signals. This is an open challenge worldwide, expected to be addressed by the development of high-fidelity EEG signals based on non-invasive electrodes and brain-machine interface technology one day. Once real EEG signal lines of human drivers are available, more precise fitting and simulation can be achieved within our provided scheme, even at the individual level with better fidelity than ever. At that time, this neuro-plausible avenue will lead towards a transcendent, stellar human-level driving understanding, just as expected by NeuroAI \cite{zador2023catalyzing}.

% 说一下我们眼动作为唯一的信息来源的可靠性
More than what we mentioned above, although vision serves as the primary information channel for active sensing while driving \cite{fang2021dada,lemonnier2020drivers,CRUNDALL2011137}, we look forward to other dimensions of sensory inputs in future work as they still play a secondary yet indispensable role in real-world driving decision-making \cite{wolfe2022toward}, such as hearing the sound of a horn and feeling the bumps on the road. Admittedly, figuring out how to integrate these sensory inputs in a neuro-plausible framework remains unanswered yet especially important for understanding the cause-effect in driving decision-making. We hope our work from a visual standpoint can open this window.

% 全文总结
Taken together, this paper investigates a commonplace yet imperative cognitive behaviour of human drivers and captures it based on a cognitively plausible model. This work is expected to bring more insightful considerations of a neuro-plausible avenue that can lead to a driving decision-making understanding. We believe that this will help us better make sense of how and why human drivers behave like that.

\section*{Methods}\label{sec4}

\subsection*{Experimental Design}\label{subsec4.1}

To understand real-world decision-making, we transform real-world driving into a trackable form. Below, we detail how to formalize real-world driving and conduct the analyses, including scenario formalization, data process, and statistical analysis.

\subsubsection*{Scenario Formalization}\label{subsubsec4.1.1}

Two types of decision-making in real-world driving are involved. Below is how we formulate them.

%% Lane-changing
The first one is \textbf{lane-changing}. Here, we discuss a general lane-changing scenario with both FV and RV driving on the nearby target lane. Some special cases, like those involving only FV or RV, can be absorbed into this general discussion. Potential interactions between the ego vehicle (EV) and both RV and FV play a decisive role in the final lane-changing decisions of human drivers. We take advantage of the fuzzy logic theory proposed in \cite{zadeh1965fuzzy} to capture these scenarios. In the framework of fuzzy logic, fuzzy sets serve as a key concept for defining multiple linguistic values to describe variables. 
This enables the utilization of imprecise language labels to describe a relative relationship between variables in situations of fuzziness or uncertainty, bypassing the need for absolute values to construct the real-world mapping. In our study, we use this theory to elaborate on the potential influence of FV and RV on EV's decision-making propensity. Three perceptual states are introduced to describe the distances of RV and FV relative to the EV: 1) near, 2) moderate, and 3) far or $\infty$ \cite{balal2016binary}. Specifically, the relative distance of RV can be represented by a fuzzy set $\widetilde{G}_\text{RV} = \{near, moderate, far/\infty\}$. Similarly, $ \widetilde{G}_\text{FV}$ is used to represent the relative distance of FV, defined as $ \widetilde{G}_\text{FV}=\{near, moderate, far/\infty\}$. Then, the fuzzy set $\widetilde{C}$ for available decisions can be defined as $\widetilde{C}=\{\textit{lane-changing}, \textit{lane-keeping}\}$.

Then, the specific value can be used to formalize the perceptual state. The perceptual state of RV is represented as $z_1$, while the perceptual state of FV is denoted as $z_2$. Their specific definitions are shown as follows.

$$z_1 = \left\{
\begin{array}{ll}
1, & \text{if } \widetilde{G}_\text{RV} \text{ is $near$} \\
2, & \text{if } \widetilde{G}_\text{RV} \text{ is $moderate$} \\
3, & \text{if } \widetilde{G}_\text{RV} \text{ is $far/\infty$}
\end{array}
\right.$$
where a lower $z_1$ value indicates the case with a relatively closer distance between EV and RV, undercutting lane-changing propensity yet favoring lane-keeping decisions of EV from a rear-view perspective. Conversely, a higher $z_1$ suggests that the state of RV is more beneficial for EV to decide lane-changing. Similarly, the value of $z_2$ represents the perceptual state of FV observed by the driver from the front view, defined by:

$$z_2 = \left\{
\begin{array}{ll}
1, & \text{if } \widetilde{G}_\text{FV} \text{ is $far/\infty$} \\
2, & \text{if } \widetilde{G}_\text{FV} \text{ is $moderate$} \\
3, & \text{if } \widetilde{G}_\text{FV} \text{ is $near$}
\end{array}
\right.$$

Based on the perceptual states of RV and FV, two evidence affordances can be developed: 1) evidence bias for lane-changing through $(z_1-z_2)=\{-2,-1,0,1,2\}$, and 2) evidence clarity through $abs(z_1 - z_2)$ as mentioned before. 
To illustrate, when $(z_1 - z_2)=2$, both RV and FV are far or $\infty$, indicating the most favorable case for lane-changing. This case has the strongest bias for changing lanes. Instead, $(z_1 - z_2)=-2$ is most favorable for lane-keeping, equaling the weakest level of bias for lane-changing. 
For evidence clarity, when $abs(z_1 - z_2)=0$, the evidence clarity is the lowest, which represents the most challenging case for a driver to make any decision. Under these conditions, drivers are provided with ambiguous evidence information relevant to decision-making.
Instead, $abs(z_1-z_2)= 2$ indicates the highest clarity, that is, a clear-to-decide case.
Taken together, the evidence affordance in lane-changing for human drivers can be formulated.

%% Car-following
Next, we can generalize this formalization to \textbf{car-following} scenarios. Unlike lane-changing, where decision-making is influenced by both FV and RV concurrently, the primary decision-making factor in car-following is the FV in the same lane. This means that the driver's attention does not need to switch between multiple traffic participants but instead focuses solely on the interplay with FV. In this context, $ \widetilde{G}_\text{FV}$ denotes the perceived state of FV by EV,  defined as $ \widetilde{G}_\text{FV}=\{\textit{FV is moving away, \textit{A moderate gap}, }\textit{FV is getting closer/braking}\}$. Next, we can reframe the perceptual state $z_1$ for FV as:

$$z_1 = \left\{
\begin{array}{ll}
1, & \text{if } \widetilde{G}_\text{FV} \text{: FV is moving away from EV} \\
2, & \text{if } \widetilde{G}_\text{FV} \text{: A moderate gap between EV and FV} \\
3, & \text{if } \widetilde{G}_\text{FV} \text{: FV is getting closer to EV / FV is braking}
\end{array}
\right.$$
where a higher value suggests that FV is approaching EV, favouring the EV driver's deceleration decisions.
Next, when the fixation is not on FV, we can define a constant perceptual state $z_2=2$ for Non-FV to represent the state of non-evidence collection from FV, which implies that the driver is unaffected or in a relaxed driving state. Following the design akin to lane-changing, evidence bias and clarity can be given by $(z_1 - z_2)$ and $abs(z_1 - z_2)$. The fuzzy set for available decisions, $\widetilde{C}$, can be defined as $\widetilde{C}=\{\textit{Decelerating}, \textit{Keep Driving}\}$. 

Collectively, both lane-changing and car-following can be formalized in a trackable manner. It is worth noting that this formalization is generalizable, offering users the flexibility to design factors related to decision-making and their corresponding perceptual states. Additional value candidates for mapping complex situations within $z_1$ and $z_2$ can be explored.

\subsubsection*{Data Processing}\label{subsubsec4.1.2}

To capture human self-evidence-collection patterns in driving, DR(eye)VE \cite{alletto2016dr}, the largest open-source real-world driving dataset that records drivers' eye movements along driving, is utilized. For lane-changing and car-following, we use two different approaches to extracting relevant latent events in which the driver has behavioural intentions.

For lane-changing extraction, we need to extract the video footage of latent lane-changing from the dataset to analyse eye movement and recognize interactive entities. %The extraction is anchored at the driver's first glance at the rear-view mirror, and a time window is predetermined to precisely extract latent lane-changing events. 
We identify segments from the entire video sequence where the driver exhibits intentions for lane-changing and their fixation alternates between FV and RV, define these as latent lane-changing events, and set a time window for each event.
Specifically, we initially record the time when the driver first glances at the rear-view mirror, denoted as $t^{1st}_\text{RV}$. Subsequently, we define the start time for decision-making $t_\text{start}$, as one second before $t^{1st}_\text{RV}$\cite{mole2021drivers}. The end time for decision-making, $t_\text{end}$, refers to the moment when the driver begins steering the wheel, or the moment when the driver no longer shifts attention. Consequently, the decision-making time window encompasses the duration from $t_\text{start}$ to $t_\text{end}$ and is utilized to analyse the driver's fixation patterns during the decision-making process of lane-changing. 
%Figure \ref{figA2}B displays the fixation patterns of drivers during the decision-making process. 
Next, for each latent lane-changing event, we employ YOLO-v8 to detect the entities of interest\cite{hussain2024yolov1,terven2023comprehensive}, including the nearest FV and the nearest RV in rear-view mirrors. As shown in Appendix Figure S1a, red rectangular bounding boxes are utilized to capture these entities existing in each lane change, yielding the positions of the target with two corner points denoted as Point1 $(x_1, y_1)$ and Point2 $(x_2, y_2)$. Effective gaze (green dot) is considered when the driver's gaze position $(x, y)$ is within the range of the recognized bounding box. As the camera position and focal length remain constant, we capture FV and RV using the bounding boxes and determine their relative distances according to the sizes of their detected bounding boxes. A larger size in the frame indicates a closer distance of FV or RV relative to EV; otherwise, a smaller size means a larger relative distance. Following the description of Section \ref{subsubsec4.1.1}, to obtain a scenario quantitative description for perceptual states of FV and RV, we recruit 13 subjects to evaluate the states of both RV and FV with $\{near, moderate, far/\infty\}$ in 30 random latent lane-changing scenarios, as shown in Figure S1b, Appendix. According to the result of their subjective experience, the average sizes of bounding boxes for FV and RV under three perceptual states are recorded as the classifying principle. Next, all latent lane-changing scenarios can be processed and scored accordingly. A real-world latent lane-changing dataset along with perceptual features can be thereby obtained. %Thus, each latent lane-changing scenario can be characterized by congruent perceptual attributes for parsing out the evidence affordance while driving. %For this analysis, we utilize the head-mounted eye-tracking glasses (ETG) video sequence from the dataset, which is captured at 720p/30fps.

For car-following extraction, we employ YOLO-v8-CLS to detect whether the EV is in car-following. We analyze the dataset to identify segments where the driver’s attention consistently shifts between FV and other non-FV areas, marking these segments as latent car-following events. 
This detection is combined with the driver’s eye movement states recorded in the DR(eye)VE dataset, which provides the labeled data of the fixation (fix on FV) and saccade (sca, fix on Non-FV) to determine the actual gaze status of the driver. 
Subsequently, we record the frequency and duration of the driver’s visual attention on FV and Non-FV, along with the interval distance between EV and FV, as well as the response behaviour of the driver. The time window for the car-following event starts one second before the first fixation on FV and ends when the driver no longer focuses on FV.
Following \ref{subsubsec4.1.1}, perceptual states are assigned based on the observed behaviour of FV. Ultimately, this process culminates in the creation of a dataset characterized by labeled values indicative of car-following behaviour. For this analysis, we utilize the roof-mounted camera video data from the dataset, which is acquired at 1080p/25fps. We project the gaze positions captured by ETG onto the video footage recorded by the roof-mounted camera, which has a significantly wider field of view. This method enhances the detection of the driver's scanning behaviour towards Non-FV during car-following, including moments when the driver briefly saccades without turning their head.

\subsubsection*{Statistical Analysis}\label{subsubsec4.1.3}

%With the feature-rich data, capturing human active sensing while driving can be conducted. 
To evaluate a model's capability to make decisions akin to human decision-making, it is essential to test whether the model generates the fundamental characteristics observed in human decision-making processes. 
For lane-changing scenarios, the extracted 139 latent lane-changing events are divided into 8 groups, we simulate an equal number of groups for the model ($N=8$), but with ten times the number of trials per group compared to the original data. For each simulated group, the trials consist of all pairwise combinations of values between 1 and 3, iterated 20 times, resulting in 180 simulation trials per simulated group, in total 1,440 simulation trials. For the car-following scenarios, we categorize the 116 latent car-following events into 6 groups. The grouping is based on the observation of more pronounced car-following deceleration behaviours on highways, contrasting with relatively fewer effective behaviours in downtown and countryside driving settings. Consequently, our analysis focuses on participant data explicitly collected from highway environments. We simulate an equal number of groups ($N=6$) with twenty times the number of trials per group compared to the original dataset, resulting in a total of 2,320 simulation trials. 

For statistical analysis, the $t$-test is employed in this paper to grasp the correlation between two decision-making evidence affordances variables (i.e., evidence bias and clarity) and decision-making behavioural features (e.g., decision probability and RT). Specifically, linear regression is adopted first to calculate the regression coefficients for each group of data. Subsequently, we utilize one-sample \textit{t}-tests to assess the significance of the regression coefficients, determining whether the relationship between decision-making mental variables and driving behavioural features is statistically significant. Finally, if the significance level \textit{p}-value is less than 0.05, 0.01, or 0.001, an analytical fact can be admitted: the linear correlation between the two variables is statistically significant. This treatment can help adjudicate the efficacy of the fitting results captured by our method. Furthermore, for the human data, the Kruskal-Wallis test is used to assess whether there is a statistically significant difference in attentional switching numbers across different levels of decision evidence clarity \cite{kruskal1952use,breslow1970generalized,chan1997learning}.

\subsection*{Model illustration}\label{subsec4.2}

Grounded in the well-documented attentional Drift Diffusion Model (aDDM), our model deliberates and incorporates numerous customizations tailored to real-world driving. Due to the consideration of evidence affordance in our scheme, our model is called the dynamic evidence accumulation model (DEAM). Below is its operational mechanism, as well as its sound cognitive plausibility. 

\subsubsection*{Dynamic Evidence Accumulation Model}\label{subsubsec4.2.1}

Under the rubric of shaping human decision-making, EAM stands out for its sound interpretability, namely, when the random motion representing the accumulation of evidence reaches the decision threshold, a decision is thus formed. The computational models derived from the specific instantiation of the EAM can be used for fitting or predicting data, which is competent in the cognitive task that involves response time and choices as key indicators \cite{ratcliff1978theory,ratcliff2016diffusion}. Consequently, some researchers advocate for the EAM as the standard method for analyzing human decision-making \cite{evans2019evidence}. Based on its framework, \cite{krajbich2010visual} enables visual consideration in the EAM by introducing an attention modulation parameter $\theta$, leading to the development of aDDM. 
However, the aDDM exhibits certain limitations, particularly in its adaptability for capturing the decision-making patterns of humans in the context of complex real-world driving scenarios.
To address this constraint, the computational model employed in this paper, termed DEAM, introduces several key advancements. First, it employs collapsing bounds instead of fixed bounds, which allows for the incorporation of increasing urgency over time in the decision-making process. Additionally, the attention modulation parameter $\theta$ is no longer constant but is dynamically adjusted based on the evidence clarity derived from each perceptual state.

Taking the lane change scenarios as an example, in each trial, $z_1$ and $z_2$ are sampled from $z_j \sim \mathcal{N}(\bar{z}, \sigma^2_z)$, $j\in \{1,2\} $ at each time step. 
In DEAM, the introduction of $z_1$ and $z_2$ represents the perceptual states corresponding to RV and FV in our study, respectively. In our problem, we assume $\bar{z_1}=\bar{z_2}$ to indicate that there is no predetermined priority for the driving decision. This is because we are investigating latent lane-changing scenarios rather than obligatory lane-changing scenarios. 
% momentary evidence
The item that the driver is attending to will provide higher information utility of evidence. Specifically, when the driver is focusing on RV, the momentary evidence received from the RV at each time step is $z_1 \Delta t$, and the momentary evidence received from FV is $\theta  \cdot z_2 \Delta t$. The attention modulation parameter $\theta$ lies within the range of 0 and 1, controlling the biasing effect of attention. A smaller $\theta$ indicates a more significant effect of attention; thus, the unattended item provides less information utility.
% RDV
In this context, DEAM can arrive at a decision by dynamically computing the relative decision value (RDV), which provides estimates of the relative attractiveness of two options at any given moment. Influenced by the evidence from the perceptual states of both RV and FV, when the driver's attention is focused on RV,  the formula for determining the RDV of the decision can be defined by
\begin{equation}
V_t =V_{t-1} + d\cdot\left({z_1} - \theta \cdot {z_2} \right) +\epsilon_t 
\end{equation}
where $V_t$ denotes the RDV at time $t$, and $d$ is a constant controlling the integration speed (see Figure S5a in the Appendix). %The attention modulation parameter $\theta$ lies within the range of 0 and 1, controlling the biasing effect of attention. 
A smaller $\theta$ indicates a greater effect of attention, thus unattended item has lower weight in the RDV accumulation process.
$\epsilon_t \sim \mathcal{N}(0, \sigma^2_\text{model})$ is white Gaussian noise with a variance of  $\sigma^2_\text{model}$.

By the same token, when the driver's attention is focused on FV,  the estimation of the RDV can be written as 
\begin{equation}
V_t =V_{t-1} + d\cdot\left({z_2} - \theta \cdot {z_1} \right) +\epsilon_t 
\end{equation}
In each trial, the attention modulation parameter $\theta$ is computed independently and negatively correlated with evidence clarity, written as
\begin{equation}
\theta  =  \frac{1}{m \cdot abs\left({z_1} -  {z_2} \right) + n}
\end{equation}
Here, the task parameters $m$ and $n$ can be adjusted for different driving tasks, allowing for flexible adaptation of the relationship between $\theta$ and evidence clarity. 
The theoretical basis of this formula suggests that in situations where evidence clarity is lower, decision-making becomes more challenging for drivers. This difficulty prompts drivers to switch attention between two items more. Therefore, in scenarios with reduced evidence clarity, the DEAM attention modulation parameter $\theta$ tends to be higher.
%The theoretical basis of this formula suggests that in situations where evidence clarity is lower, drivers find it more difficult for decision-making. This tendency encourages drivers to switch attention between two items more. Therefore, in scenarios with lower evidence clarity, the DEAM attention modulation parameter $\theta$ tends to be higher.

%边界的公式：
According to the EAM model, decision-makers continuously accumulate evidence until they reach a decision bound. In DEAM, the decision bound is a collapse bound, whose absolute value decreases over time from its initial value ($B_\text{start}$) to an asymptotic value. We describe the bound using an exponential function, where parameter $r$ defines the speed of bound collapse (Figure S5b, Appendix). The upper bound is defined as 
\begin{equation}
B_\text{upper}(t) = e^{- r \cdot t} \cdot B_\text{start}
\end{equation}
and the lower bound is
\begin{equation}
B_\text{lower}(t) = -  e^{- r \cdot t} \cdot B_\text{start}
\end{equation}
When RDV reaches either bound, decision-makers cease further accumulation of evidence and finalize their decision-making. 

Considering the six free parameters in DEAM: $d$, $m$, $n$, $r$, $B_\text{start}$ and $\sigma$, we need to find a group of parameters with sound performance. Thus, a genetic algorithm (GA) is adopted to fit the key parameters of DEAM. In closing, the adopted parameters for lane-changing scenarios are $d=0.003$, $m = 0.18$, $n = 1.25$, $r=0.35$, $B_\text{start} = 2.8$, and  $\sigma=0.03$. The adopted DEAM parameters for car-following scenarios are $d= 0.0008$, $m = 0.1$, $n = 1.5$, $r=0.15$, $B_\text{start} = 1.5$ and  $\sigma=0.01$.
 
% To incorporate the utility of evidence information, we begin by assuming that the overall rate of evidence from both items, controlled by $\sigma^2$, remains constant. Then, attention parameter $\theta$ regulates the proportionate amount of information acquired about the attended item compared to the unattended item. This implies that for the scenario the driver is attending to, the variance of the momentary evidence for the item is  $\theta \cdot \sigma_x^2 \delta t$, while for the non-attended item, the variance of the momentary evidence is $\sigma_x^2 \delta t$. Consequently, the unattended item's information variance is greater than that of the attended item, making the attended one more informative. The total variance of the information provided by the evidence from all items is represented by $\sigma^2 \delta t$.

\subsubsection*{Cognitive Plausibility}
For DEAM in shaping human decision-making, some inherent theoretical assumptions should be satisfied to meet cognitive plausibility, including six principles \cite{liubehavioral}: universality of evidence accumulation, evidence accumulating selectively, attention-driven modulation of evidence, linear integration of evidence with noise, collapsing decision criteria, and independence of decision-making and motor processes. Below is how these principles are contextualized in our problem:

\begin{enumerate}

    \item The universality of evidence accumulation: Humans can accumulate evidence for decision-making from various sources of information regardless of specific conditions or scenarios \cite{o2012supramodal}. In driving scenarios, this information is mainly perceptual, i.e., the evidence from vision, which can be regarded as descriptive information for the individual understanding of driving situations. The perceptual states of FV, RV, and Non-FV in our study do respond to this principle, and two kinds of evidence affordances are developed accordingly. 
    
    \item Evidence accumulation selectivity \cite{zilker2022stronger,loughnane2016target,krajbich2010visual}: This assumption defines which stimuli can be encoded as evidence. Humans selectively ignore information irrelevant to the task goal and only accumulate decision-relevant information as evidence. In driving scenarios, drivers selectively transform the decision-relevant information, such as driving reachability and behaviour of social vehicles, into usable evidence. In lane-changing scenarios, the FV perceived from the front windshield view and the RV from the rear-view mirror decide the evidence sources jointly. For car-following scenarios, the afforded evidence depends solely on the state of FV.

    \item Attentional modulation of evidence: This assumption describes the modulatory effect of attention on decision-making, where attention modulates the amount of momentary evidence information about the attended and unattended items \cite{jang2021optimal,tajima2019optimal,tavares2017attentional,krajbich2010visual}. Attention limits the evidence information of unattended items, thereby giving greater weight to evidence accumulation related to the currently attended item in the decision-making process. 
    
    \item Linear integration of evidence with noise \cite{cochrane2023multiple,stine2020differentiating,roitman2002response,kelly2013internal}: This assumption describes the way decision-relevant evidence is integrated. Generally, drivers linearly integrate information from the external road scene, taking into account the variability and uncertainty present in the observed scene. This assumption acknowledges that the evidence accumulation is influenced by noise $\epsilon_t$, resulting in the decisions of drivers exhibiting some level of randomness.
    
    \item Collapsing decision-making criteria: Traditional diffusion models assume that humans maintain a consistent decision criterion across different trials \cite{hawkins2015revisiting,voss2019sequential,stine2023neural}. However, a newer and more complex assumption gaining popularity involves collapsing decision-making bounds \cite{palestro2018some,drugowitsch2012cost,bowman2012temporal,thura2012decision,cisek2009decisions}. These collapsing bounds, sometimes seen as urgency signals, allow decisions to be triggered with progressively less evidence accumulation as time elapses \cite{kelly2021neurocomputational}. Collapsing bounds can reduce the occurrence of slow decision-making, which is particularly relevant for scenarios such as driving where humans typically need to make rapid decisions. %Figure ** illustrates how models with  consistent and collapsing bounds yield different predictions for response times

    \item Independence of decision-making process and behaviour \cite{callaway2021fixation,gluth2018value,o2012supramodal}: Decision-making refers to the process where drivers accumulate evidence and decide whether to engage in intended driving behaviours. Behaviour, on the other hand, refers to the execution of the corresponding action, such as steering maneuver and deceleration, after the decision is formed. These two processes occur sequentially and independently, and the behaviour process does not interfere with the decision-making process. 
 
\end{enumerate}

%\newpage

%\bmhead{Supplementary information}

%\bmhead{Acknowledgments}

\section*{Declarations}

\begin{itemize}
%\item Funding
\item Competing interests\\
The authors declare no competing interests.

%\item Ethics approval 
%\item Consent to participate
%\item Consent for publication
\item Availability of data\\
Original data can be found at \url{http://imagelab.ing.unimore.it/dreyeve}

 \item Code availability \\
For any questions regarding code availability, please contact yliuii@connect.ust.hk

\end{itemize}

%\begin{appendices}

%\section{Extended figures}\label{secA1}

%\begin{figure}[H]
%\centering
%\includegraphics[width=0.75\textwidth]{dataprocess.png}
%\caption{(a) Data processing by YOLO. (b) Scenario evaluation by subjects.}\label{figS1}
%\end{figure}

%%=============================================%%
%% For submissions to Nature Portfolio Journals %%
%% please use the heading ``Extended Data''.   %%
%%=============================================%%

%%=============================================================%%
%% Sample for another appendix section			       %%
%%=============================================================%%

%% \section{Example of another appendix section}\label{secA2}%
%% Appendices may be used for helpful, supporting or essential material that would otherwise 
%% clutter, break up or be distracting to the text. Appendices can consist of sections, figures, 
%% tables and equations etc.

%\end{appendices}

%%===========================================================================================%%
%% If you are submitting to one of the Nature Portfolio journals, using the eJP submission   %%
%% system, please include the references within the manuscript file itself. You may do this  %%
%% by copying the reference list from your .bbl file, paste it into the main manuscript .tex %%
%% file, and delete the associated \verb+\bibliography+ commands.                            %%
%%===========================================================================================%%
\newpage
\bibliography{sn-bibliography}% common bib file

\begin{thebibliography}{10}
\urlstyle{rm}
\expandafter\ifx\csname url\endcsname\relax
  \def\url#1{\texttt{#1}}\fi
\expandafter\ifx\csname urlprefix\endcsname\relax\def\urlprefix{URL }\fi
\expandafter\ifx\csname doiprefix\endcsname\relax\def\doiprefix{DOI: }\fi
\providecommand{\bibinfo}[2]{#2}
\providecommand{\eprint}[2][]{\url{#2}}

\bibitem{thomas2019gaze}
\bibinfo{author}{Thomas, A.~W.}, \bibinfo{author}{Molter, F.}, \bibinfo{author}{Krajbich, I.}, \bibinfo{author}{Heekeren, H.~R.} \& \bibinfo{author}{Mohr, P.~N.}
\newblock \bibinfo{journal}{\bibinfo{title}{Gaze bias differences capture individual choice behaviour}}.
\newblock {\emph{\JournalTitle{Nature Human Behaviour}}} \textbf{\bibinfo{volume}{3}}, \bibinfo{pages}{625--635} (\bibinfo{year}{2019}).

\bibitem{gottlieb2012attention}
\bibinfo{author}{Gottlieb, J.}
\newblock \bibinfo{journal}{\bibinfo{title}{Attention, learning, and the value of information}}.
\newblock {\emph{\JournalTitle{Neuron}}} \textbf{\bibinfo{volume}{76}}, \bibinfo{pages}{281--295} (\bibinfo{year}{2012}).

\bibitem{gottlieb2018towards}
\bibinfo{author}{Gottlieb, J.} \& \bibinfo{author}{Oudeyer, P.-Y.}
\newblock \bibinfo{journal}{\bibinfo{title}{Towards a neuroscience of active sampling and curiosity}}.
\newblock {\emph{\JournalTitle{Nature Reviews Neuroscience}}} \textbf{\bibinfo{volume}{19}}, \bibinfo{pages}{758--770} (\bibinfo{year}{2018}).

\bibitem{yang2016theoretical}
\bibinfo{author}{Yang, S. C.-H.}, \bibinfo{author}{Wolpert, D.~M.} \& \bibinfo{author}{Lengyel, M.}
\newblock \bibinfo{journal}{\bibinfo{title}{Theoretical perspectives on active sensing}}.
\newblock {\emph{\JournalTitle{Current opinion in behavioral sciences}}} \textbf{\bibinfo{volume}{11}}, \bibinfo{pages}{100--108} (\bibinfo{year}{2016}).

\bibitem{vellenga2022driver}
\bibinfo{author}{Vellenga, K.} \emph{et~al.}
\newblock \bibinfo{journal}{\bibinfo{title}{Driver intention recognition: State-of-the-art review}}.
\newblock {\emph{\JournalTitle{IEEE Open Journal of Intelligent Transportation Systems}}} \textbf{\bibinfo{volume}{3}}, \bibinfo{pages}{602--616} (\bibinfo{year}{2022}).

\bibitem{blakemore2001perception}
\bibinfo{author}{Blakemore, S.-J.} \& \bibinfo{author}{Decety, J.}
\newblock \bibinfo{journal}{\bibinfo{title}{From the perception of action to the understanding of intention}}.
\newblock {\emph{\JournalTitle{Nature reviews neuroscience}}} \textbf{\bibinfo{volume}{2}}, \bibinfo{pages}{561--567} (\bibinfo{year}{2001}).

\bibitem{glickman2022evidence}
\bibinfo{author}{Glickman, M.}, \bibinfo{author}{Moran, R.} \& \bibinfo{author}{Usher, M.}
\newblock \bibinfo{journal}{\bibinfo{title}{Evidence integration and decision confidence are modulated by stimulus consistency}}.
\newblock {\emph{\JournalTitle{Nature Human Behaviour}}} \textbf{\bibinfo{volume}{6}}, \bibinfo{pages}{988--999} (\bibinfo{year}{2022}).

\bibitem{hoxha2023accounting}
\bibinfo{author}{Hoxha, I.} \emph{et~al.}
\newblock \bibinfo{journal}{\bibinfo{title}{Accounting for endogenous effects in decision-making with a non-linear diffusion decision model}}.
\newblock {\emph{\JournalTitle{Scientific Reports}}} \textbf{\bibinfo{volume}{13}}, \bibinfo{pages}{6323} (\bibinfo{year}{2023}).

\bibitem{van2019relation}
\bibinfo{author}{van Vugt, M.~K.}, \bibinfo{author}{Beulen, M.~A.} \& \bibinfo{author}{Taatgen, N.~A.}
\newblock \bibinfo{journal}{\bibinfo{title}{Relation between centro-parietal positivity and diffusion model parameters in both perceptual and memory-based decision making}}.
\newblock {\emph{\JournalTitle{Brain research}}} \textbf{\bibinfo{volume}{1715}}, \bibinfo{pages}{1--12} (\bibinfo{year}{2019}).

\bibitem{kiverstein2024experience}
\bibinfo{author}{Kiverstein, J.} \& \bibinfo{author}{Artese, G.~F.}
\newblock \bibinfo{journal}{\bibinfo{title}{The experience of affordances in an intersubjective world}}.
\newblock {\emph{\JournalTitle{Topoi}}} \textbf{\bibinfo{volume}{43}}, \bibinfo{pages}{187--200} (\bibinfo{year}{2024}).

\bibitem{jones2003affordance}
\bibinfo{author}{Jones, K.~S.}
\newblock \bibinfo{journal}{\bibinfo{title}{What is an affordance?}}
\newblock {\emph{\JournalTitle{Ecological Psychology}}} \textbf{\bibinfo{volume}{15}}, \bibinfo{pages}{107--114} (\bibinfo{year}{2003}).

\bibitem{gibson1977theory}
\bibinfo{author}{Gibson, J.~J.}
\newblock \bibinfo{journal}{\bibinfo{title}{The theory of affordances}}.
\newblock {\emph{\JournalTitle{Hilldale, USA}}} \textbf{\bibinfo{volume}{1}}, \bibinfo{pages}{67--82} (\bibinfo{year}{1977}).

\bibitem{sepulveda2020visual}
\bibinfo{author}{Sepulveda, P.} \emph{et~al.}
\newblock \bibinfo{journal}{\bibinfo{title}{Visual attention modulates the integration of goal-relevant evidence and not value}}.
\newblock {\emph{\JournalTitle{Elife}}} \textbf{\bibinfo{volume}{9}}, \bibinfo{pages}{e60705} (\bibinfo{year}{2020}).

\bibitem{shevlin2021attention}
\bibinfo{author}{Shevlin, B.~R.} \& \bibinfo{author}{Krajbich, I.}
\newblock \bibinfo{journal}{\bibinfo{title}{Attention as a source of variability in decision-making: Accounting for overall-value effects with diffusion models}}.
\newblock {\emph{\JournalTitle{Journal of Mathematical Psychology}}} \textbf{\bibinfo{volume}{105}}, \bibinfo{pages}{102594} (\bibinfo{year}{2021}).

\bibitem{callaway2021fixation}
\bibinfo{author}{Callaway, F.}, \bibinfo{author}{Rangel, A.} \& \bibinfo{author}{Griffiths, T.~L.}
\newblock \bibinfo{journal}{\bibinfo{title}{Fixation patterns in simple choice reflect optimal information sampling}}.
\newblock {\emph{\JournalTitle{PLoS computational biology}}} \textbf{\bibinfo{volume}{17}}, \bibinfo{pages}{e1008863} (\bibinfo{year}{2021}).

\bibitem{rangelov2020evidence}
\bibinfo{author}{Rangelov, D.} \& \bibinfo{author}{Mattingley, J.~B.}
\newblock \bibinfo{journal}{\bibinfo{title}{Evidence accumulation during perceptual decision-making is sensitive to the dynamics of attentional selection}}.
\newblock {\emph{\JournalTitle{NeuroImage}}} \textbf{\bibinfo{volume}{220}}, \bibinfo{pages}{117093} (\bibinfo{year}{2020}).

\bibitem{gottlieb2014attention}
\bibinfo{author}{Gottlieb, J.}, \bibinfo{author}{Hayhoe, M.}, \bibinfo{author}{Hikosaka, O.} \& \bibinfo{author}{Rangel, A.}
\newblock \bibinfo{journal}{\bibinfo{title}{Attention, reward, and information seeking}}.
\newblock {\emph{\JournalTitle{Journal of Neuroscience}}} \textbf{\bibinfo{volume}{34}}, \bibinfo{pages}{15497--15504} (\bibinfo{year}{2014}).

\bibitem{orquin2013attention}
\bibinfo{author}{Orquin, J.~L.} \& \bibinfo{author}{Loose, S.~M.}
\newblock \bibinfo{journal}{\bibinfo{title}{Attention and choice: A review on eye movements in decision making}}.
\newblock {\emph{\JournalTitle{Acta psychologica}}} \textbf{\bibinfo{volume}{144}}, \bibinfo{pages}{190--206} (\bibinfo{year}{2013}).

\bibitem{ahlstrom2021eye}
\bibinfo{author}{Ahlstr{\"o}m, C.}, \bibinfo{author}{Kircher, K.}, \bibinfo{author}{Nystr{\"o}m, M.} \& \bibinfo{author}{Wolfe, B.}
\newblock \bibinfo{journal}{\bibinfo{title}{Eye tracking in driver attention research—how gaze data interpretations influence what we learn}}.
\newblock {\emph{\JournalTitle{Frontiers in Neuroergonomics}}} \textbf{\bibinfo{volume}{2}}, \bibinfo{pages}{778043} (\bibinfo{year}{2021}).

\bibitem{alletto2016dr}
\bibinfo{author}{Alletto, S.}, \bibinfo{author}{Palazzi, A.}, \bibinfo{author}{Solera, F.}, \bibinfo{author}{Calderara, S.} \& \bibinfo{author}{Cucchiara, R.}
\newblock \bibinfo{title}{Dr (eye) ve: a dataset for attention-based tasks with applications to autonomous and assisted driving}.
\newblock In \emph{\bibinfo{booktitle}{Proceedings of the ieee conference on computer vision and pattern recognition workshops}}, \bibinfo{pages}{54--60} (\bibinfo{year}{2016}).

\bibitem{palazzi2018predicting}
\bibinfo{author}{Palazzi, A.}, \bibinfo{author}{Abati, D.}, \bibinfo{author}{Solera, F.}, \bibinfo{author}{Cucchiara, R.} \emph{et~al.}
\newblock \bibinfo{journal}{\bibinfo{title}{Predicting the driver's focus of attention: the dr (eye) ve project}}.
\newblock {\emph{\JournalTitle{IEEE transactions on pattern analysis and machine intelligence}}} \textbf{\bibinfo{volume}{41}}, \bibinfo{pages}{1720--1733} (\bibinfo{year}{2018}).

\bibitem{harkins1987information}
\bibinfo{author}{Harkins, S.~G.} \& \bibinfo{author}{Petty, R.~E.}
\newblock \bibinfo{journal}{\bibinfo{title}{Information utility and the multiple source effect.}}
\newblock {\emph{\JournalTitle{Journal of personality and social psychology}}} \textbf{\bibinfo{volume}{52}}, \bibinfo{pages}{260} (\bibinfo{year}{1987}).

\bibitem{stewart2022humans}
\bibinfo{author}{Stewart, E.~E.}, \bibinfo{author}{Ludwig, C.~J.} \& \bibinfo{author}{Sch{\"u}tz, A.~C.}
\newblock \bibinfo{journal}{\bibinfo{title}{Humans represent the precision and utility of information acquired across fixations}}.
\newblock {\emph{\JournalTitle{Scientific Reports}}} \textbf{\bibinfo{volume}{12}}, \bibinfo{pages}{2411} (\bibinfo{year}{2022}).

\bibitem{gluth2020value}
\bibinfo{author}{Gluth, S.}, \bibinfo{author}{Kern, N.}, \bibinfo{author}{Kortmann, M.} \& \bibinfo{author}{Vitali, C.~L.}
\newblock \bibinfo{journal}{\bibinfo{title}{Value-based attention but not divisive normalization influences decisions with multiple alternatives}}.
\newblock {\emph{\JournalTitle{Nature human behaviour}}} \textbf{\bibinfo{volume}{4}}, \bibinfo{pages}{634--645} (\bibinfo{year}{2020}).

\bibitem{wedel2023modeling}
\bibinfo{author}{Wedel, M.}, \bibinfo{author}{Pieters, R.} \& \bibinfo{author}{van~der Lans, R.}
\newblock \bibinfo{journal}{\bibinfo{title}{Modeling eye movements during decision making: A review}}.
\newblock {\emph{\JournalTitle{Psychometrika}}} \textbf{\bibinfo{volume}{88}}, \bibinfo{pages}{697--729} (\bibinfo{year}{2023}).

\bibitem{zhou2009cognitive}
\bibinfo{author}{Zhou, H.}, \bibinfo{author}{Itoh, M.} \& \bibinfo{author}{Inagaki, T.}
\newblock \bibinfo{title}{How do cognitive distraction affect driver intent of changing lanes?}
\newblock In \emph{\bibinfo{booktitle}{Intelligent Robotics and Applications: Second International Conference, ICIRA 2009, Singapore, December 16-18, 2009. Proceedings 2}}, \bibinfo{pages}{235--244} (\bibinfo{organization}{Springer}, \bibinfo{year}{2009}).

\bibitem{jang2021optimal}
\bibinfo{author}{Jang, A.~I.}, \bibinfo{author}{Sharma, R.} \& \bibinfo{author}{Drugowitsch, J.}
\newblock \bibinfo{journal}{\bibinfo{title}{Optimal policy for attention-modulated decisions explains human fixation behavior}}.
\newblock {\emph{\JournalTitle{Elife}}} \textbf{\bibinfo{volume}{10}}, \bibinfo{pages}{e63436} (\bibinfo{year}{2021}).

\bibitem{krajbich2010visual}
\bibinfo{author}{Krajbich, I.}, \bibinfo{author}{Armel, C.} \& \bibinfo{author}{Rangel, A.}
\newblock \bibinfo{journal}{\bibinfo{title}{Visual fixations and the computation and comparison of value in simple choice}}.
\newblock {\emph{\JournalTitle{Nature neuroscience}}} \textbf{\bibinfo{volume}{13}}, \bibinfo{pages}{1292--1298} (\bibinfo{year}{2010}).

\bibitem{tavares2017attentional}
\bibinfo{author}{Tavares, G.}, \bibinfo{author}{Perona, P.} \& \bibinfo{author}{Rangel, A.}
\newblock \bibinfo{journal}{\bibinfo{title}{The attentional drift diffusion model of simple perceptual decision-making}}.
\newblock {\emph{\JournalTitle{Frontiers in neuroscience}}} \textbf{\bibinfo{volume}{11}}, \bibinfo{pages}{468} (\bibinfo{year}{2017}).

\bibitem{generalization2022}
\bibinfo{author}{Editorial}.
\newblock \bibinfo{journal}{\bibinfo{title}{Replication studies hold the key to generalization}}.
\newblock {\emph{\JournalTitle{Nature Communications}}} \textbf{\bibinfo{volume}{13}} (\bibinfo{year}{2022}).

\bibitem{baee2021medirl}
\bibinfo{author}{Baee, S.} \emph{et~al.}
\newblock \bibinfo{title}{Medirl: Predicting the visual attention of drivers via maximum entropy deep inverse reinforcement learning}.
\newblock In \emph{\bibinfo{booktitle}{Proceedings of the IEEE/CVF international conference on computer vision}}, \bibinfo{pages}{13178--13188} (\bibinfo{year}{2021}).

\bibitem{Gottlieb_Oudeyer_Lopes_Baranes_2013}
\bibinfo{author}{Gottlieb, J.}, \bibinfo{author}{Oudeyer, P.-Y.}, \bibinfo{author}{Lopes, M.} \& \bibinfo{author}{Baranes, A.}
\newblock \bibinfo{journal}{\bibinfo{title}{Information-seeking, curiosity, and attention: computational and neural mechanisms}}.
\newblock {\emph{\JournalTitle{Trends in Cognitive Sciences}}} \bibinfo{pages}{585–593} (\bibinfo{year}{2013}).

\bibitem{gottlieb2020curiosity}
\bibinfo{author}{Gottlieb, J.}, \bibinfo{author}{Cohanpour, M.}, \bibinfo{author}{Li, Y.}, \bibinfo{author}{Singletary, N.} \& \bibinfo{author}{Zabeh, E.}
\newblock \bibinfo{journal}{\bibinfo{title}{Curiosity, information demand and attentional priority}}.
\newblock {\emph{\JournalTitle{Current Opinion in Behavioral Sciences}}} \textbf{\bibinfo{volume}{35}}, \bibinfo{pages}{83--91} (\bibinfo{year}{2020}).

\bibitem{milosavljevic2010drift}
\bibinfo{author}{Milosavljevic, M.}, \bibinfo{author}{Malmaud, J.}, \bibinfo{author}{Huth, A.}, \bibinfo{author}{Koch, C.} \& \bibinfo{author}{Rangel, A.}
\newblock \bibinfo{journal}{\bibinfo{title}{The drift diffusion model can account for the accuracy and reaction time of value-based choices under high and low time pressure}}.
\newblock {\emph{\JournalTitle{Judgment and Decision making}}} \textbf{\bibinfo{volume}{5}}, \bibinfo{pages}{437--449} (\bibinfo{year}{2010}).

\bibitem{tajima2016optimal}
\bibinfo{author}{Tajima, S.}, \bibinfo{author}{Drugowitsch, J.} \& \bibinfo{author}{Pouget, A.}
\newblock \bibinfo{journal}{\bibinfo{title}{Optimal policy for value-based decision-making}}.
\newblock {\emph{\JournalTitle{Nature communications}}} \textbf{\bibinfo{volume}{7}}, \bibinfo{pages}{12400} (\bibinfo{year}{2016}).

\bibitem{kobayashi2019common}
\bibinfo{author}{Kobayashi, K.} \& \bibinfo{author}{Hsu, M.}
\newblock \bibinfo{journal}{\bibinfo{title}{Common neural code for reward and information value}}.
\newblock {\emph{\JournalTitle{Proceedings of the National Academy of Sciences}}} \textbf{\bibinfo{volume}{116}}, \bibinfo{pages}{13061--13066} (\bibinfo{year}{2019}).

\bibitem{kobayashi2024neural}
\bibinfo{author}{Kobayashi, K.} \& \bibinfo{author}{Kable, J.~W.}
\newblock \bibinfo{journal}{\bibinfo{title}{Neural mechanisms of information seeking}}.
\newblock {\emph{\JournalTitle{Neuron}}}  (\bibinfo{year}{2024}).

\bibitem{bobadilla2020subjective}
\bibinfo{author}{Bobadilla-Suarez, S.}, \bibinfo{author}{Guest, O.} \& \bibinfo{author}{Love, B.~C.}
\newblock \bibinfo{journal}{\bibinfo{title}{Subjective value and decision entropy are jointly encoded by aligned gradients across the human brain}}.
\newblock {\emph{\JournalTitle{Communications biology}}} \textbf{\bibinfo{volume}{3}}, \bibinfo{pages}{597} (\bibinfo{year}{2020}).

\bibitem{caplin2001psychological}
\bibinfo{author}{Caplin, A.} \& \bibinfo{author}{Leahy, J.}
\newblock \bibinfo{journal}{\bibinfo{title}{Psychological expected utility theory and anticipatory feelings}}.
\newblock {\emph{\JournalTitle{The Quarterly Journal of Economics}}} \textbf{\bibinfo{volume}{116}}, \bibinfo{pages}{55--79} (\bibinfo{year}{2001}).

\bibitem{marques2020internal}
\bibinfo{author}{Marques, J.~C.}, \bibinfo{author}{Li, M.}, \bibinfo{author}{Schaak, D.}, \bibinfo{author}{Robson, D.~N.} \& \bibinfo{author}{Li, J.~M.}
\newblock \bibinfo{journal}{\bibinfo{title}{Internal state dynamics shape brainwide activity and foraging behaviour}}.
\newblock {\emph{\JournalTitle{Nature}}} \textbf{\bibinfo{volume}{577}}, \bibinfo{pages}{239--243} (\bibinfo{year}{2020}).

\bibitem{sharot2020people}
\bibinfo{author}{Sharot, T.} \& \bibinfo{author}{Sunstein, C.~R.}
\newblock \bibinfo{journal}{\bibinfo{title}{How people decide what they want to know}}.
\newblock {\emph{\JournalTitle{Nature Human Behaviour}}} \textbf{\bibinfo{volume}{4}}, \bibinfo{pages}{14--19} (\bibinfo{year}{2020}).

\bibitem{kobayashi2019diverse}
\bibinfo{author}{Kobayashi, K.}, \bibinfo{author}{Ravaioli, S.}, \bibinfo{author}{Baran{\`e}s, A.}, \bibinfo{author}{Woodford, M.} \& \bibinfo{author}{Gottlieb, J.}
\newblock \bibinfo{journal}{\bibinfo{title}{Diverse motives for human curiosity}}.
\newblock {\emph{\JournalTitle{Nature human behaviour}}} \textbf{\bibinfo{volume}{3}}, \bibinfo{pages}{587--595} (\bibinfo{year}{2019}).

\bibitem{nunez2017attention}
\bibinfo{author}{Nunez, M.~D.}, \bibinfo{author}{Vandekerckhove, J.} \& \bibinfo{author}{Srinivasan, R.}
\newblock \bibinfo{journal}{\bibinfo{title}{How attention influences perceptual decision making: Single-trial eeg correlates of drift-diffusion model parameters}}.
\newblock {\emph{\JournalTitle{Journal of mathematical psychology}}} \textbf{\bibinfo{volume}{76}}, \bibinfo{pages}{117--130} (\bibinfo{year}{2017}).

\bibitem{si2020predicting}
\bibinfo{author}{Si, Y.} \emph{et~al.}
\newblock \bibinfo{journal}{\bibinfo{title}{Predicting individual decision-making responses based on single-trial eeg}}.
\newblock {\emph{\JournalTitle{NeuroImage}}} \textbf{\bibinfo{volume}{206}}, \bibinfo{pages}{116333} (\bibinfo{year}{2020}).

\bibitem{zador2023catalyzing}
\bibinfo{author}{Zador, A.} \emph{et~al.}
\newblock \bibinfo{journal}{\bibinfo{title}{Catalyzing next-generation artificial intelligence through neuroai}}.
\newblock {\emph{\JournalTitle{Nature communications}}} \textbf{\bibinfo{volume}{14}}, \bibinfo{pages}{1597} (\bibinfo{year}{2023}).

\bibitem{fang2021dada}
\bibinfo{author}{Fang, J.}, \bibinfo{author}{Yan, D.}, \bibinfo{author}{Qiao, J.}, \bibinfo{author}{Xue, J.} \& \bibinfo{author}{Yu, H.}
\newblock \bibinfo{journal}{\bibinfo{title}{Dada: Driver attention prediction in driving accident scenarios}}.
\newblock {\emph{\JournalTitle{IEEE transactions on intelligent transportation systems}}} \textbf{\bibinfo{volume}{23}}, \bibinfo{pages}{4959--4971} (\bibinfo{year}{2021}).

\bibitem{lemonnier2020drivers}
\bibinfo{author}{Lemonnier, S.}, \bibinfo{author}{D{\'e}sir{\'e}, L.}, \bibinfo{author}{Bremond, R.} \& \bibinfo{author}{Baccino, T.}
\newblock \bibinfo{journal}{\bibinfo{title}{Drivers’ visual attention: A field study at intersections}}.
\newblock {\emph{\JournalTitle{Transportation research part F: traffic psychology and behaviour}}} \textbf{\bibinfo{volume}{69}}, \bibinfo{pages}{206--221} (\bibinfo{year}{2020}).

\bibitem{CRUNDALL2011137}
\bibinfo{author}{Crundall, D.} \& \bibinfo{author}{Underwood, G.}
\newblock \bibinfo{title}{Chapter 11 - visual attention while driving: Measures of eye movements used in driving research}.
\newblock In \bibinfo{editor}{Porter, B.~E.} (ed.) \emph{\bibinfo{booktitle}{Handbook of Traffic Psychology}}, \bibinfo{pages}{137--148} (\bibinfo{publisher}{Academic Press}, \bibinfo{address}{San Diego}, \bibinfo{year}{2011}).

\bibitem{wolfe2022toward}
\bibinfo{author}{Wolfe, B.}, \bibinfo{author}{Sawyer, B.~D.} \& \bibinfo{author}{Rosenholtz, R.}
\newblock \bibinfo{journal}{\bibinfo{title}{Toward a theory of visual information acquisition in driving}}.
\newblock {\emph{\JournalTitle{Human factors}}} \textbf{\bibinfo{volume}{64}}, \bibinfo{pages}{694--713} (\bibinfo{year}{2022}).

\bibitem{zadeh1965fuzzy}
\bibinfo{author}{Zadeh, L.~A.}
\newblock \bibinfo{journal}{\bibinfo{title}{Fuzzy sets}}.
\newblock {\emph{\JournalTitle{Information and control}}} \textbf{\bibinfo{volume}{8}}, \bibinfo{pages}{338--353} (\bibinfo{year}{1965}).

\bibitem{balal2016binary}
\bibinfo{author}{Balal, E.}, \bibinfo{author}{Cheu, R.~L.} \& \bibinfo{author}{Sarkodie-Gyan, T.}
\newblock \bibinfo{journal}{\bibinfo{title}{A binary decision model for discretionary lane changing move based on fuzzy inference system}}.
\newblock {\emph{\JournalTitle{Transportation Research Part C: Emerging Technologies}}} \textbf{\bibinfo{volume}{67}}, \bibinfo{pages}{47--61} (\bibinfo{year}{2016}).

\bibitem{mole2021drivers}
\bibinfo{author}{Mole, C.}, \bibinfo{author}{Pekkanen, J.}, \bibinfo{author}{Sheppard, W.~E.}, \bibinfo{author}{Markkula, G.} \& \bibinfo{author}{Wilkie, R.~M.}
\newblock \bibinfo{journal}{\bibinfo{title}{Drivers use active gaze to monitor waypoints during automated driving}}.
\newblock {\emph{\JournalTitle{Scientific reports}}} \textbf{\bibinfo{volume}{11}}, \bibinfo{pages}{263} (\bibinfo{year}{2021}).

\bibitem{hussain2024yolov1}
\bibinfo{author}{Hussain, M.}
\newblock \bibinfo{journal}{\bibinfo{title}{Yolov1 to v8: Unveiling each variant--a comprehensive review of yolo}}.
\newblock {\emph{\JournalTitle{IEEE Access}}} \textbf{\bibinfo{volume}{12}}, \bibinfo{pages}{42816--42833} (\bibinfo{year}{2024}).

\bibitem{terven2023comprehensive}
\bibinfo{author}{Terven, J.}, \bibinfo{author}{C{\'o}rdova-Esparza, D.-M.} \& \bibinfo{author}{Romero-Gonz{\'a}lez, J.-A.}
\newblock \bibinfo{journal}{\bibinfo{title}{A comprehensive review of yolo architectures in computer vision: From yolov1 to yolov8 and yolo-nas}}.
\newblock {\emph{\JournalTitle{Machine Learning and Knowledge Extraction}}} \textbf{\bibinfo{volume}{5}}, \bibinfo{pages}{1680--1716} (\bibinfo{year}{2023}).

\bibitem{kruskal1952use}
\bibinfo{author}{Kruskal, W.~H.} \& \bibinfo{author}{Wallis, W.~A.}
\newblock \bibinfo{journal}{\bibinfo{title}{Use of ranks in one-criterion variance analysis}}.
\newblock {\emph{\JournalTitle{Journal of the American statistical Association}}} \textbf{\bibinfo{volume}{47}}, \bibinfo{pages}{583--621} (\bibinfo{year}{1952}).

\bibitem{breslow1970generalized}
\bibinfo{author}{Breslow, N.}
\newblock \bibinfo{journal}{\bibinfo{title}{A generalized kruskal-wallis test for comparing k samples subject to unequal patterns of censorship}}.
\newblock {\emph{\JournalTitle{Biometrika}}} \textbf{\bibinfo{volume}{57}}, \bibinfo{pages}{579--594} (\bibinfo{year}{1970}).

\bibitem{chan1997learning}
\bibinfo{author}{Chan, Y.} \& \bibinfo{author}{Walmsley, R.~P.}
\newblock \bibinfo{journal}{\bibinfo{title}{Learning and understanding the kruskal-wallis one-way analysis-of-variance-by-ranks test for differences among three or more independent groups}}.
\newblock {\emph{\JournalTitle{Physical therapy}}} \textbf{\bibinfo{volume}{77}}, \bibinfo{pages}{1755--1761} (\bibinfo{year}{1997}).

\bibitem{ratcliff1978theory}
\bibinfo{author}{Ratcliff, R.}
\newblock \bibinfo{journal}{\bibinfo{title}{A theory of memory retrieval.}}
\newblock {\emph{\JournalTitle{Psychological review}}} \textbf{\bibinfo{volume}{85}}, \bibinfo{pages}{59} (\bibinfo{year}{1978}).

\bibitem{ratcliff2016diffusion}
\bibinfo{author}{Ratcliff, R.}, \bibinfo{author}{Smith, P.~L.}, \bibinfo{author}{Brown, S.~D.} \& \bibinfo{author}{McKoon, G.}
\newblock \bibinfo{journal}{\bibinfo{title}{Diffusion decision model: Current issues and history}}.
\newblock {\emph{\JournalTitle{Trends in cognitive sciences}}} \textbf{\bibinfo{volume}{20}}, \bibinfo{pages}{260--281} (\bibinfo{year}{2016}).

\bibitem{evans2019evidence}
\bibinfo{author}{Evans, N.~J.} \& \bibinfo{author}{Wagenmakers, E.-J.}
\newblock \bibinfo{title}{Evidence accumulation models: Current limitations and future directions}.
\newblock \bibinfo{howpublished}{PsyArXiv}, \url{10.31234/osf.io/xyz12} (\bibinfo{year}{2019}).

\bibitem{liubehavioral}
\bibinfo{author}{Liu, Y.} \& \bibinfo{author}{Hu, C.-P.}
\newblock \bibinfo{journal}{\bibinfo{title}{Behavioral and cognitive neuroscience findings regarding assumptions of the evidence accumulation model}}.
\newblock {\emph{\JournalTitle{Science Bulletin}}} \url{10.1360/TB-2023-1080} (\bibinfo{year}{2024}).

\bibitem{o2012supramodal}
\bibinfo{author}{O'connell, R.~G.}, \bibinfo{author}{Dockree, P.~M.} \& \bibinfo{author}{Kelly, S.~P.}
\newblock \bibinfo{journal}{\bibinfo{title}{A supramodal accumulation-to-bound signal that determines perceptual decisions in humans}}.
\newblock {\emph{\JournalTitle{Nature neuroscience}}} \textbf{\bibinfo{volume}{15}}, \bibinfo{pages}{1729--1735} (\bibinfo{year}{2012}).

\bibitem{zilker2022stronger}
\bibinfo{author}{Zilker, V.}
\newblock \bibinfo{journal}{\bibinfo{title}{Stronger attentional biases can be linked to higher reward rate in preferential choice}}.
\newblock {\emph{\JournalTitle{Cognition}}} \textbf{\bibinfo{volume}{225}}, \bibinfo{pages}{105095} (\bibinfo{year}{2022}).

\bibitem{loughnane2016target}
\bibinfo{author}{Loughnane, G.~M.} \emph{et~al.}
\newblock \bibinfo{journal}{\bibinfo{title}{Target selection signals influence perceptual decisions by modulating the onset and rate of evidence accumulation}}.
\newblock {\emph{\JournalTitle{Current Biology}}} \textbf{\bibinfo{volume}{26}}, \bibinfo{pages}{496--502} (\bibinfo{year}{2016}).

\bibitem{tajima2019optimal}
\bibinfo{author}{Tajima, S.}, \bibinfo{author}{Drugowitsch, J.}, \bibinfo{author}{Patel, N.} \& \bibinfo{author}{Pouget, A.}
\newblock \bibinfo{journal}{\bibinfo{title}{Optimal policy for multi-alternative decisions}}.
\newblock {\emph{\JournalTitle{Nature neuroscience}}} \textbf{\bibinfo{volume}{22}}, \bibinfo{pages}{1503--1511} (\bibinfo{year}{2019}).

\bibitem{cochrane2023multiple}
\bibinfo{author}{Cochrane, A.}, \bibinfo{author}{Sims, C.~R.}, \bibinfo{author}{Bejjanki, V.~R.}, \bibinfo{author}{Green, C.~S.} \& \bibinfo{author}{Bavelier, D.}
\newblock \bibinfo{journal}{\bibinfo{title}{Multiple timescales of learning indicated by changes in evidence-accumulation processes during perceptual decision-making}}.
\newblock {\emph{\JournalTitle{npj Science of Learning}}} \textbf{\bibinfo{volume}{8}}, \bibinfo{pages}{19} (\bibinfo{year}{2023}).

\bibitem{stine2020differentiating}
\bibinfo{author}{Stine, G.~M.}, \bibinfo{author}{Zylberberg, A.}, \bibinfo{author}{Ditterich, J.} \& \bibinfo{author}{Shadlen, M.~N.}
\newblock \bibinfo{journal}{\bibinfo{title}{Differentiating between integration and non-integration strategies in perceptual decision making}}.
\newblock {\emph{\JournalTitle{Elife}}} \textbf{\bibinfo{volume}{9}}, \bibinfo{pages}{e55365} (\bibinfo{year}{2020}).

\bibitem{roitman2002response}
\bibinfo{author}{Roitman, J.~D.} \& \bibinfo{author}{Shadlen, M.~N.}
\newblock \bibinfo{journal}{\bibinfo{title}{Response of neurons in the lateral intraparietal area during a combined visual discrimination reaction time task}}.
\newblock {\emph{\JournalTitle{Journal of neuroscience}}} \textbf{\bibinfo{volume}{22}}, \bibinfo{pages}{9475--9489} (\bibinfo{year}{2002}).

\bibitem{kelly2013internal}
\bibinfo{author}{Kelly, S.~P.} \& \bibinfo{author}{O'Connell, R.~G.}
\newblock \bibinfo{journal}{\bibinfo{title}{Internal and external influences on the rate of sensory evidence accumulation in the human brain}}.
\newblock {\emph{\JournalTitle{Journal of Neuroscience}}} \textbf{\bibinfo{volume}{33}}, \bibinfo{pages}{19434--19441} (\bibinfo{year}{2013}).

\bibitem{hawkins2015revisiting}
\bibinfo{author}{Hawkins, G.~E.}, \bibinfo{author}{Forstmann, B.~U.}, \bibinfo{author}{Wagenmakers, E.-J.}, \bibinfo{author}{Ratcliff, R.} \& \bibinfo{author}{Brown, S.~D.}
\newblock \bibinfo{journal}{\bibinfo{title}{Revisiting the evidence for collapsing boundaries and urgency signals in perceptual decision-making}}.
\newblock {\emph{\JournalTitle{Journal of Neuroscience}}} \textbf{\bibinfo{volume}{35}}, \bibinfo{pages}{2476--2484} (\bibinfo{year}{2015}).

\bibitem{voss2019sequential}
\bibinfo{author}{Voss, A.}, \bibinfo{author}{Lerche, V.}, \bibinfo{author}{Mertens, U.} \& \bibinfo{author}{Voss, J.}
\newblock \bibinfo{journal}{\bibinfo{title}{Sequential sampling models with variable boundaries and non-normal noise: A comparison of six models}}.
\newblock {\emph{\JournalTitle{Psychonomic bulletin \& review}}} \textbf{\bibinfo{volume}{26}}, \bibinfo{pages}{813--832} (\bibinfo{year}{2019}).

\bibitem{stine2023neural}
\bibinfo{author}{Stine, G.~M.}, \bibinfo{author}{Trautmann, E.~M.}, \bibinfo{author}{Jeurissen, D.} \& \bibinfo{author}{Shadlen, M.~N.}
\newblock \bibinfo{journal}{\bibinfo{title}{A neural mechanism for terminating decisions}}.
\newblock {\emph{\JournalTitle{Neuron}}}  (\bibinfo{year}{2023}).

\bibitem{palestro2018some}
\bibinfo{author}{Palestro, J.~J.}, \bibinfo{author}{Weichart, E.}, \bibinfo{author}{Sederberg, P.~B.} \& \bibinfo{author}{Turner, B.~M.}
\newblock \bibinfo{journal}{\bibinfo{title}{Some task demands induce collapsing bounds: Evidence from a behavioral analysis}}.
\newblock {\emph{\JournalTitle{Psychonomic bulletin \& review}}} \textbf{\bibinfo{volume}{25}}, \bibinfo{pages}{1225--1248} (\bibinfo{year}{2018}).

\bibitem{drugowitsch2012cost}
\bibinfo{author}{Drugowitsch, J.}, \bibinfo{author}{Moreno-Bote, R.}, \bibinfo{author}{Churchland, A.~K.}, \bibinfo{author}{Shadlen, M.~N.} \& \bibinfo{author}{Pouget, A.}
\newblock \bibinfo{journal}{\bibinfo{title}{The cost of accumulating evidence in perceptual decision making}}.
\newblock {\emph{\JournalTitle{Journal of Neuroscience}}} \textbf{\bibinfo{volume}{32}}, \bibinfo{pages}{3612--3628} (\bibinfo{year}{2012}).

\bibitem{bowman2012temporal}
\bibinfo{author}{Bowman, N.~E.}, \bibinfo{author}{Kording, K.~P.} \& \bibinfo{author}{Gottfried, J.~A.}
\newblock \bibinfo{journal}{\bibinfo{title}{Temporal integration of olfactory perceptual evidence in human orbitofrontal cortex}}.
\newblock {\emph{\JournalTitle{Neuron}}} \textbf{\bibinfo{volume}{75}}, \bibinfo{pages}{916--927} (\bibinfo{year}{2012}).

\bibitem{thura2012decision}
\bibinfo{author}{Thura, D.}, \bibinfo{author}{Beauregard-Racine, J.}, \bibinfo{author}{Fradet, C.-W.} \& \bibinfo{author}{Cisek, P.}
\newblock \bibinfo{journal}{\bibinfo{title}{Decision making by urgency gating: theory and experimental support}}.
\newblock {\emph{\JournalTitle{Journal of neurophysiology}}} \textbf{\bibinfo{volume}{108}}, \bibinfo{pages}{2912--2930} (\bibinfo{year}{2012}).

\bibitem{cisek2009decisions}
\bibinfo{author}{Cisek, P.}, \bibinfo{author}{Puskas, G.~A.} \& \bibinfo{author}{El-Murr, S.}
\newblock \bibinfo{journal}{\bibinfo{title}{Decisions in changing conditions: the urgency-gating model}}.
\newblock {\emph{\JournalTitle{Journal of Neuroscience}}} \textbf{\bibinfo{volume}{29}}, \bibinfo{pages}{11560--11571} (\bibinfo{year}{2009}).

\bibitem{kelly2021neurocomputational}
\bibinfo{author}{Kelly, S.~P.}, \bibinfo{author}{Corbett, E.~A.} \& \bibinfo{author}{O’Connell, R.~G.}
\newblock \bibinfo{journal}{\bibinfo{title}{Neurocomputational mechanisms of prior-informed perceptual decision-making in humans}}.
\newblock {\emph{\JournalTitle{Nature Human Behaviour}}} \textbf{\bibinfo{volume}{5}}, \bibinfo{pages}{467--481} (\bibinfo{year}{2021}).

\bibitem{gluth2018value}
\bibinfo{author}{Gluth, S.}, \bibinfo{author}{Spektor, M.~S.} \& \bibinfo{author}{Rieskamp, J.}
\newblock \bibinfo{journal}{\bibinfo{title}{Value-based attentional capture affects multi-alternative decision making}}.
\newblock {\emph{\JournalTitle{Elife}}} \textbf{\bibinfo{volume}{7}}, \bibinfo{pages}{e39659} (\bibinfo{year}{2018}).

\end{thebibliography}
%% If required, the content of .bbl file can be included here once bbl is generated
%%\input sn-article.bbl

\end{document}